\pdfoutput=1

\documentclass[11pt,table]{article}

\usepackage[final]{acl}
\usepackage{amsthm}
\usepackage{times}
\usepackage{latexsym}
\usepackage{amsmath}
\usepackage{amssymb, geometry}
\usepackage[T1]{fontenc}
\usepackage{booktabs} 
\usepackage{multirow}
\usepackage{adjustbox}
\usepackage{float, graphicx}

\usepackage{titlesec}
\titlespacing{\paragraph}{0pt}{5pt}{2pt}
\usepackage[utf8]{inputenc}
\usepackage{bbm}
\usepackage{xcolor}
\usepackage{hyperref}
\usepackage{microtype}

\usepackage{inconsolata}

\usepackage{graphicx}
\usepackage{algpseudocode}
\usepackage[ruled,linesnumbered]{algorithm2e}

\newcommand{\gate}{\textrm{gate}}
\usepackage{stmaryrd}
\title{Bridging Robustness and Generalization Against Word Substitution Attacks in NLP via the Growth Bound Matrix Approach}

\author{
  Mohammed Bouri\textsuperscript{1,2} \and
  Adnane Saoud\textsuperscript{1} \\
  \textsuperscript{1}College of Computing, Mohammed VI Polytechnic University, Morocco \\
  \textsuperscript{2}CID Development, Morocco \\
  \texttt{mohammed.bouri@um6p.ma}, \texttt{adnane.saoud@um6p.ma}
}

\begin{document}
\maketitle
\begin{abstract}
Despite advancements in Natural Language Processing (NLP), models remain vulnerable to adversarial attacks, such as synonym substitutions. While prior work has focused on improving robustness for feed-forward and convolutional architectures, the robustness of recurrent networks and modern state space models (SSMs), such as S4, remains understudied. These architectures pose unique challenges due to their sequential processing and complex parameter dynamics. In this paper, we introduce a novel regularization technique based on Growth Bound Matrices (GBM) to improve NLP model robustness by reducing the impact of input perturbations on model outputs. We focus on computing the GBM for three architectures: Long Short-Term Memory (LSTM), State Space models (S4), and Convolutional Neural Networks (CNN). Our method aims to (1) enhance resilience against word substitution attacks, (2) improve generalization on clean text, and (3) providing the first systematic analysis of SSM (S4) robustness. Extensive experiments across multiple architectures and benchmark datasets demonstrate that our method improves adversarial robustness by up to \(8.8\%\) over existing baselines. These results highlight the effectiveness of our approach, outperforming several state-of-the-art methods in adversarial defense. {Codes are available at \href{https://github.com/BouriMohammed/GBM}{https://github.com/BouriMohammed/GBM}}
\end{abstract}

\section{Introduction}

\begin{figure}[t!]
    \centering
    \includegraphics[width=1.\columnwidth]{ 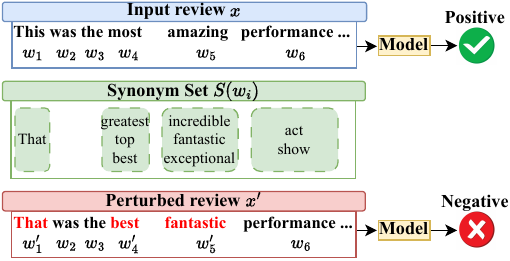}
        \vspace{-.2in}
    \caption{{Word substitution-based perturbations in sentiment analysis. The original sentence $x$ (top) and its perturbed version $x'$ (bottom) differ by synonym replacements $w_i \to w'_i$ from $S(w_i)$. Despite preserving meaning, these changes can alter the model’s predicted sentiment (e.g., Positive to Negative) due to its sensitivity to specific word choices.}
}
    \label{fig:title_overview}
    \vspace{-.2in}
\end{figure}
Deep learning models have demonstrated impressive results across various machine learning tasks, particularly in Language Modeling (LM), which has provided powerful tools for Natural Language Processing (NLP). However, recent studies have shown that these models are vulnerable to adversarial examples. Initially, adversarial examples were identified in image classification tasks \citep{goodfellow2014explaining, qi2024exploring}, leading to a surge of research focused on adversarial attacks in NLP tasks. These adversarial examples pose significant challenges for real-world applications such as text classification \citep{song2021universal}. {Additionally, although Large Language Models (LLMs) are commonly employed as decoders for text generation, they are also susceptible to adversarial examples when used in text classification tasks. \citep{wang2024generating, shayegani2023survey, yang2024assessing}.}\\
Textual adversarial attacks can be categorized into three types: character-level perturbations, sentence-level attacks, and word-level attacks. Character-level perturbations \citep{javid18hotflip,eger2020hero} can often be mitigated by spell checkers \citep{danish19combating}. Sentence-level attacks \citep{wang2019t3,pei2022generating} utilize paraphrasing but typically fail to preserve the original sentence semantics. Word-level attacks \citep{shuhuai19generating,alzantot-etal-2018-generating,yuan20pso,wang21adversarial,rishabh21generating}, which rely on synonym substitutions, have become the most widely adopted approach. These attacks can craft adversarial examples with high success rates while maintaining grammatical correctness and semantic consistency, making them particularly challenging to defend against, as shown in Figure (\ref{fig:title_overview}). Consequently, our work focuses on introducing a new robust training method against word-level substitution adversarial attacks.\\
To enhance robustness against such attacks, several studies have developed defense methods by generating adversarial examples and integrating them into the training set \citep{shuhuai19generating, alzantot-etal-2018-generating}. However, generating enough adversarial examples per epoch is computationally expensive due to the discrete nature of the input space. To accelerate adversarial training, \citep{wang21adversarial} and \citep{dong21towards} introduced improvements by designing fast white-box adversarial attacks. Despite these advancements, the computation time remains significantly longer than standard training. 
In another scope, certified defense methods based on interval bound propagation (IBP)~\citep{jia19certified, huang19achieving} offer theoretical lower bounds on robustness. Nevertheless, these methods are computationally intensive, and demonstrate an overly conservative robustness. \citep{zhang-etal-2021-certified} introduced a novel approach, Abstractive Recursive Certification (ARC), which defines a set of programmatically perturbed string transformations and constructs a perturbation space using these transformations. The perturbation sets are represented as a hyperrectangle, which is then propagated through the network using the (IBP) technique. However, this approach is limited in its effectiveness to two-word substitutions; increasing the number of substitutions significantly degrades the performance.\\
In this work, we introduce Growth Bound Matrices (GBM) as a novel method to enhance the robustness of NLP models. GBM provides certified robustness against perturbations, enabling models to maintain generalization while reducing sensitivity to input variations. Furthermore, we present a mathematical framework for computing GBM across different architectures, including Long Short-Term Memory (LSTM) networks \citep{hochreiter1997long}, State Space Model S4 \citep{GuGR22}, and Convolutional Neural Networks (CNN) \citep{Kim14f}.
Extensive experiments on multiple benchmark datasets reveal that GBM significantly enhances the certified robust accuracy of models. For instance, on the IMDB dataset, GBM achieves an impressive $84.3\%$ certified robust accuracy, surpassing IBP by approximately $16.7\%$. \\
Our main contributions are summarized as follows:
\begin{itemize}
    \item \textbf{Certified robustness with GBM:} We introduce Growth Bound Matrices (GBM), a novel framework that establishes certifiable robustness guarantees by bounding input-output variations.
    \vspace{-1.5mm}
    \item \textbf{Empirical validation across models and datasets:} Extensive experiments demonstrate that minimizing GBM significantly enhances adversarial robustness across multiple NLP models and benchmark datasets. Moreover, our approach improves adversarial robustness over existing defense baselines.
    \vspace{-1.3mm}
    \item \textbf{First study on SSM robustness in NLP:} To the best of our knowledge, this is the first work to investigate the generalization and robustness of the State Space Model (SSM) S4 in text classification.
\end{itemize}

\section{Related Work}



Spelling and grammar checkers have been shown to be robust against character-level and sentence-level attacks, which often violate grammatical requirements \citep{pruthi-etal-2019-combating, ge-etal-2019-automatic}. However, these methods are ineffective against word-level attacks.\\
Existing defense methods against word-level attacks can be categorized as follows:\\ 
Adversarial training (AT) is one of the most popular defense strategies \citep{goodfellow2014explaining, madry18towards, alzantot-etal-2018-generating, shuhuai19generating, ivgi21achieving, zhu2019freelb, li2020tavat, wang2020infobert, zhou2020defense}. Based on the Fast Gradient Projection Method (FGPM), an adversarial text attack technique, Adversarial Training with FGPM enhanced by Logit Pairing (ATFL) \citep{wang21adversarial} generates adversarial examples and injects them into the training set. Region-based adversarial training \citep{zhou2020defense, dong21towards} enhances model robustness by optimizing performance within the convex hull of a word’s embedding and its synonyms.\\
Certified defense methods aim to ensure robustness against all adversarial perturbations \citep{zeng21certified, wang21certified, huang19achieving}. Interval Bound Propagation (IBP)-based methods \citep{jia19certified} certify robustness by iteratively computing upper and lower bounds of the output for an interval-constrained input, minimizing the worst-case loss from word substitutions. However, IBP methods struggle with scalability due to high computational costs and strict constraints. Randomized smoothing-based methods \citep{zhang2024text, yeetal2020safer, zeng2021certified} are structure-free, constructing stochastic ensembles of input texts and leveraging their statistical properties to certify robustness. Moreover, most certified defense methods assume access to the attacker's synonym set, which is unrealistic, as adversaries are not restricted in their choice of synonyms.

\section{Preliminaries}
\label{subsec:Preliminaries}
In this paper, we address the text classification problem. Consider a model \(f: \mathcal{X} \rightarrow \mathcal{Y}\) that predicts a label \(y \in \mathcal{Y}\) for a given input \(x \in \mathcal{X}\), where \(x = \langle w_1, w_2, \ldots, w_N \rangle\) is a sequence of \(N\) words, where each word \(w_i \in \mathcal{W}\), and \(\mathcal{W}\) denotes the vocabulary set. The output space \(\mathcal{Y} = \{y_1, y_2, \ldots, y_c\}\) denotes all classification labels. Standard NLP models embed the input $x$ into a sequence of vectors \(v_x = \langle v_{w_1}, v_{w_2}, \ldots, v_{w_N} \rangle\) and classify via \(p(y|v_x)\), parameterized using a neural network.\\
Adversarial robustness against synonym substitutions has been extensively studied in recent literature~\citep{alzantot-etal-2018-generating, jia19certified, wang21natural, dong21towards, zhang2024text}. The synonym set for a given word \(w_i\), denoted as \(\mathcal{S}(w_i)\), is constructed by selecting the \(k\) nearest words to \(w_i\) within an Euclidean distance {\(d_e\)} in the embedding space. 
{To ensure semantic consistency, the perturbed input space is defined as follows:}
\begin{equation*}
    \mathcal{S}_{adv}(x) = \{\langle w_1', \ldots, w_N' \rangle \mid w_i' \in \mathcal{S}(w_i) \cup \{w_i\}\}.
\end{equation*}
Our adversarial defense aims to certify the robustness of the model by ensuring that:
\begin{equation*}
    \forall x^{\prime} \in \mathcal{S}_{adv}(x), \quad f(x^{\prime}) = f(x) = y.
\end{equation*}
The objective of this work is to ensure that for all possible combinations of words in a sentence, when replaced by their synonyms, the classification outcome remains consistent, as shown in Figure \ref{fig:title_overview}.

\section{Methodology}
\label{sec:Methodology}

In this section, we present our main contribution. We first introduce the concept of Growth Bound Matrices (GBM) and explore how integrating these matrices into the training process can enhance model robustness. We then provide a detailed computation of GBMs for multiple architectures, including LSTM networks, State Space Model S4 and CNN models.

\subsection{Growth Bound Matrix (GBM)}

The Growth Bound Matrix (GBM)-based method is designed to quantify and control model sensitivity to input perturbations. Specifically, GBM establishes theoretical guarantees on the maximum output variation of a model \( \mathcal{F} \) when its input \( x \) is subjected to a small perturbation \( {\delta} \).

\subsubsection{Definition and Theoretical Foundation}
Consider a mapping, 
\begin{equation}\label{eq:orig}
\begin{array}{l}
\mathcal{F} :\hspace{2.5mm} \mathbf{X} \to \mathbf{Y} \\
\quad\quad\hspace{2mm} x \mapsto \mathcal{F}(x)
\end{array}
\end{equation}
where \(\mathbf{X} \subseteq \mathbb{R}^{n_x}\) and \(\mathbf{Y} \subseteq \mathbb{R}^{n_y}\).
\paragraph{Definition (Growth Bound Matrix):} A matrix \( \mathcal{M} \in \mathbb{R}^{n_y \times n_x} \) is said to be a \textit{Growth Bound Matrix (GBM)} for the mapping \( \mathcal{F} \) in (\ref{eq:orig}) if the following condition holds:
\begin{equation}\label{eq:GBM}
    \left\lVert \dfrac{\partial \mathcal{F}^i}{\partial x^j}(x)\right\rVert \leq (\mathcal{M})_{ij} \quad \forall (i,j) \in \mathcal{I}, \quad \forall x \in \mathbf{X}
\end{equation}
where \( \mathcal{I} = \{1,\ldots,n_y\} \times \{1,\ldots,n_x\} \) is a predefined index set, and \( (\mathcal{M})_{ij} \) denotes the entry in the \( i \)-th row and \( j \)-th column of \( \mathcal{M} \).

The matrix \( \mathcal{M}\) capture the element-wise maximum of the boundaries of the partial derivative. By evaluating this GBM, we gain insights into the sensitivity of the output of a model with respect to its input. Therefore, reducing the GBM enables us to improve the robustness of the considered models.

\subsubsection{Robustness via GBM}

\begin{figure*}[ht]
\vspace{-.1in}
\centering
\includegraphics[width=0.95\textwidth]{ 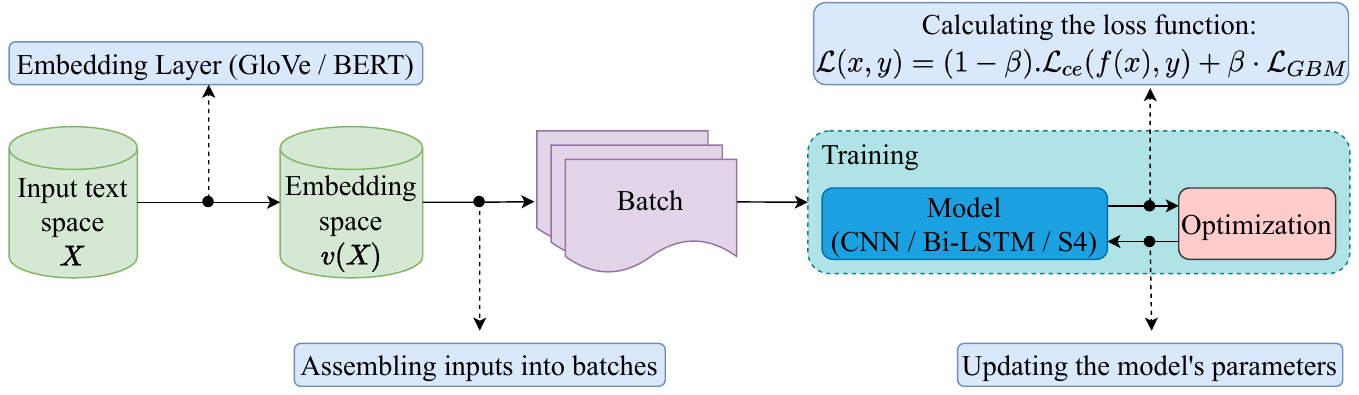}
\caption{Training Pipeline with Growth Bound Matrix (GBM) Regularization. This pipeline integrates GBM regularization into the training process of a deep learning model. First, the input text \( X \) is converted into a vector representation \( v(X) \) using an embedding layer (e.g., GloVe or BERT). The inputs are grouped into batches, and the loss function \( \mathcal{L}(x, y) \) is calculated as a weighted sum of the cross-entropy loss \( \mathcal{L}_{ce} \) and the GBM regularization term \( \mathcal{L}_{GBM} \). The model's parameters (e.g., CNN, BiLSTM, or S4) are then updated while minimizing the GBM term to enhance robustness against adversarial perturbations.}\label{fig:overview}
\vspace{-.1in}
\end{figure*}
In this section, we formally demonstrate how the GBM provides certified robustness. The proof is provided in Appendix \ref{sec:RGBM}.

\paragraph{Proposition 1:} 
\label{sub:prop1}
Consider the mapping \( \mathcal{F} \) in (\ref{eq:orig}) and let a matrix \( \mathcal{M} \in \mathbb{R}^{n_y \times n_x} \) be its GBM. Given an input \( x \in \mathbf{X} \), a perturbation vector \( {\delta} \in \mathbb{R}^{n_x} \) and consider the perturbed input $x'\in \mathbf{X}$ such that $x'=x+{\delta}$. Then, for each component $\mathcal{F}_i$, $i\in \{1,\ldots,n_y\}$ , the following inequality holds:
\setlength{\abovedisplayskip}{7pt}
\begin{equation*}
    \begin{aligned}
          \mathcal{F}^i(x) - \sum\limits_{j=1}^{n_x}(\mathcal{M})_{ij}\|{\delta_j}\| &\leq \mathcal{F}^i(x')\\[-5mm]
          &\hspace{-2mm}\leq \mathcal{F}^i(x) + \sum\limits_{j=1}^{n_x}(\mathcal{M})_{ij}\|{\delta_j}\|
    \end{aligned}
\end{equation*}
where \({\delta_j}\) denotes the \(j\)-th component of the perturbation vector ${\delta}$.

{This result indicates that the change in the ith component of the output \(\mathcal{F}^i\) caused by the perturbation \({\delta}\) is bounded by the corresponding row of \(\mathcal{M}\). By minimizing the GBM \(\mathcal{M}\), we effectively reduce the sensitivity of the function \( \mathcal{F} \) to perturbations in \( x \). This provides a certification that the outputs of the perturbed and non-perturbed inputs remain within the same class, ensuring robustness against small input variations.}

\subsection{Overall Training Objective}

{The aim is to train a model that is consistent with the data while minimizing the GBM.} To achieve this, we incorporate a regularization term into the neural network’s loss function, thereby enhancing the model’s robustness against attacks or perturbations. {An illustration of the proposed training pipeline is given in Fig.~\ref{fig:overview}.} We formulate the overall training objective as follows:
\begin{equation}
    \label{eq:objective}
    \mathcal{L}(x,y)= (1-\beta).\mathcal{L}_{ce}(f(x), y) + \beta \cdot \mathcal{L}_{GBM},
\end{equation}
where
\setlength{\abovedisplayskip}{5pt}
\setlength{\belowdisplayskip}{5pt}
\[
\mathcal{L}_{GBM}= \sum_{i=1}^{n_y}\sum_{j=1}^{n_x} (\mathcal{M})_{ij} \quad.
\]
Here, \( \mathcal{L}_{ce}(\cdot,\cdot) \) denotes the cross-entropy loss function employed to train the layers following the embedding layer, primarily focusing on classification accuracy. The matrix \( \mathcal{M} \) represents the GBM introduced in Eq.~\eqref{eq:GBM}, and its specific formulations for each model (Eqs.~\eqref{eq:GBMLSTM}, \eqref{eq:GBMS4}, \eqref{eq:GBMcnn}) are detailed in Subsection~\ref{subsec:models}. The term \( \beta \) serves as a hyperparameter that balances model accuracy and robustness. Meanwhile, the regularization term \( \mathcal{L}_{GBM} \) aims to minimize the elements of \( \mathcal{M} \), thereby enhancing the model’s resilience to adversarial attacks. By including this regularization component in the training objective (Fig.~\ref{fig:overview}), the resulting model exhibits improved robustness.

\subsection{Model-Specific GBM}
\label{subsec:models}

In this paper, we derive model-specific GBMs for three architectures: LSTM, S4, and CNN.

\paragraph{Long Short-Term Memory (LSTM) Network}

LSTM networks are an improved variant of Recurrent Neural Networks (RNNs). LSTMs were developed to address the vanishing and exploding gradient problems that occur during the training of traditional RNNs \citep{hochreiter1997long}. \\
Given a sentence of \(N\) input word vectors \(v_x = \langle v_{w_1}, v_{w_2}, \ldots,v_{w_N} \rangle\), where \(v_{w_t} \in \mathbb{R}^{d_0}\) represents the embedding of the word at position \( t \) in the sentence.
We begin by defining the LSTM cell model, which is characterized by the cell state $c_t \in \mathbb{R}^d$ and the hidden state \(h_t \in \mathbb{R}^d\). These states are updated according to the following equations:
\begin{align}
    c_t &= f_t \odot c_{t-1} + I_t \odot g_t \label{eq:ct} \\
    h_t &= o_t \odot \tanh\left(c_t\right) \label{eq:ht}
\end{align}
where $\odot$ denotes element-wise multiplication, $I_t$ is the input gate, $f_t$ is the forget gate, $g_t$ is the cell gate, and $o_t$ is the output gate, each gate governed by its own set of learnable parameters (weights, and biases). These gates control the flow of information through the network. {Based on Eqs.\ (\ref{eq:ct}) and (\ref{eq:ht}), we can represent the LSTM cell as an input--output mapping \(\mathcal{F}\), as in Eq.\ (\ref{eq:orig}), formally defined as:
\begin{multline}\label{eq:lstm}
\mathcal{F} : \mathbf{V} \times \mathbf{H} \times \mathbf{C} \to \mathbb{R}^d \\
(v_{w_t}, h_{t-1}, c_{t-1}) \mapsto o_t \odot \tanh(f_t \odot c_{t-1} + I_t \odot g_t)
\end{multline}
where the input \(\ x=(v_{w_t}, h_{t-1}, c_{t-1}) \in \mathbf{V} \times \mathbf{H} \times \mathbf{C} \subseteq \mathbb{R}^{d_0+2d}\) and the output \(\,y = h_t = \mathcal{F}(v_{w_t}, h_{t-1}, c_{t-1}) \in \mathbb{R}^d\). Here, \(\mathbf{V}\) denotes the domain of the word vectors, \(\mathbf{H}\) denotes the domain of hidden states, and \(\mathbf{C}\) denotes the domain of cell states.
}\\{The following result provide an explicit formula for the GBM of the LSTM cell. A detailed proof of this result, as well as the algorithm used to compute the GBM, can be found in the Appendix (\ref{sec:proofs},\ref{sec:Algo}). }

\paragraph{Proposition 2:}
\label{prop2}
{Consider the map $\mathcal{F}$ describing the input-output model of an LSTM cell defined in Eq.\eqref{eq:lstm}. The GBM of the map $\mathcal{F}$ can be expressed as follows: 
\[\text{For all }(i,j) \in  \{1,\ldots,d\} \times \{1,\ldots,d_0+2d\} \]
\begin{equation}\label{eq:GBMLSTM}
    (\mathcal{M})_{ij} = \max\left(\| (\underline{\mathcal{M}})_{ij} \|, \| (\overline{\mathcal{M}})_{ij} \|\right),
\end{equation}
where, for all input \(x=(v_{w_t}, h_{t-1}, c_{t-1}) \in \mathbf{X}= (\mathbf{V} \times \mathbf{H} \times \mathbf{C})\)
\begin{small}
    \begin{equation*}
(\underline{\mathcal{M}})_{i,j} =
\begin{cases}
      \min\{\hspace{1mm}({\mathcal{M}}_v)_{i,j} (x)
\mid x \in \mathbf{X}\} \hspace{2mm} \text{if} \hspace{2mm} j\in \mathcal{J}_{1}\\
     \min\{\hspace{1mm}({\mathcal{M}}_h)_{i,j-d_0} (x)
\mid x \in \mathbf{X}\}\hspace{1mm} \text{if} \hspace{1mm} j\in \mathcal{J}_{2}\\
     \min\{\hspace{1mm}({\mathcal{M}}_c)_{i,j-d_0-d} (x)
\mid x \in \mathbf{X}\}\hspace{1mm} \text{if} \hspace{1mm} j\in \mathcal{J}_{3}
\end{cases}
\end{equation*}
\end{small}
\begin{small}
    \begin{equation*}
(\overline{\mathcal{M}})_{i,j} =
\begin{cases}
      \max\{\hspace{1mm}({\mathcal{M}}_v)_{i,j} (x)
\mid x \in \mathbf{X}\} \hspace{1mm} \text{if} \hspace{1mm} j\in \mathcal{J}_{1}\\
     \max\{\hspace{1mm}({\mathcal{M}}_h)_{i,j-d_0} (x)
\mid x \in \mathbf{X}\}\hspace{1mm} \text{if} \hspace{1mm} j\in \mathcal{J}_{2}\\
     \max\{\hspace{1mm}({\mathcal{M}}_c)_{i,j-d_0-d} (x)
\mid x \in \mathbf{X}\}\hspace{1mm} \text{if} \hspace{1mm} j\in \mathcal{J}_{3}
\end{cases}
\end{equation*}
\end{small}
with \(\mathcal{J}_{1} = \{1,\ldots,d_0\}\), \(\mathcal{J}_{2} = \{d_0+1,\ldots,d_0+d\}\) and \(\mathcal{J}_{3} = \{d_0+d+1,\ldots,d_0+2d\}\)\\
where the map \(x \mapsto ({\mathcal{M}}_v)_{i,j} (x)\) is given by:
\[
\begin{aligned}
    (\mathcal{M}_v)_{i,j}(x) &= (\Theta^{(o)})_{ij} .\sigma'(T_o^i) .\tanh(c^i_{t}) \\
    &+ \sigma(T_o^i) .\frac{\partial c^i_t}{\partial v_{w_t}^j} .\tanh'(c^i_{t}),
\end{aligned}
\]
the map \(x \mapsto ({\mathcal{M}}_h)_{i,j} (x)\) is given by:
\[
\begin{aligned}
    (\mathcal{M}_h)_{i,j}(x) &= (U^{(o)})_{ij} .\sigma'(T_o^i) .\tanh(c^i_{t}) \\
    &+ \sigma(T_o^i) .\frac{\partial c^i_t}{\partial h_{t-1}^j} .\tanh'(c^i_{t}),
\end{aligned}
\]
the map \(x \mapsto ({\mathcal{M}}_c)_{i,j} (x)\) is given by:
\[
\begin{aligned}
   (\mathcal{M}_c)_{i,j} (x)&=\sigma(T_o^i).\sigma(T_f^i).\tanh'(c^i_{t})
\end{aligned}
\]
and the map \(x \mapsto c^i_t (x)\) is given by:
\[
\begin{aligned}
   c^i_t &= f^i_t \odot c^i_{t-1} + I_t^i \odot g^i_t
\end{aligned}
\]
with:
\begin{small}
    \begin{equation*}
    T_{\gate}^i = \sum_{p=1}^{d_0} (\Theta^{(\gate)})_{ip}.v_{w_t}^{(p)} + \sum_{q=1}^{d} (U^{(\gate)})_{iq}.h_{t-1}^{(q)} + b_i^{(\gate)}
\end{equation*}
\end{small}
Here, $\sigma$ denotes the sigmoid function and $\tanh$ denotes the hyperbolic tangent function, \(\sigma'\) and \(\tanh'\) denote their respective derivatives. For $\gate \in \{f,o\}$, the parameters $\Theta^{(gate)} \in \mathbb{R}^{d\times d_0} , U^{(gate)} \in \mathbb{R}^{d\times d}$ and $b^{(gate)} \in \mathbb{R}^{d}$ are the input-hidden weights, hidden-hidden weights, and biases, respectively.}
\hspace{-4mm}\textbf{State Space Model (SSM) S4}\\
The S4 has demonstrated outstanding performance in text classification tasks, surpassing transformers in various benchmarks \citep{GuGR22}.\\
{Given a sentence represented as a sequence of \( N \) word embeddings, where each word at position \( t \) is associated with a vector \( v_{w_t} \in \mathbb{R}^{d_0} \). Collectively, these embeddings form the input sequence \( v_x = \langle v_{w_1}, v_{w_2}, \dots, v_{w_N} \rangle \).}\\
The continuous-time State Space Model (SSM) is designed to process each dimension of the input independently while utilizing shared parameters across all dimensions. The continuous-time formulation of the SSM is given by:
\begin{align*}
    \dot{h}(t) &= A h(t) + B v_{w_t}, \\
    y(t) &= C h(t) + D v_{w_t},
\end{align*}
where \(h(t) \in \mathbb{R}^{d_0d}\) represents the hidden state at time \(t\). The parameters \(A \in \mathbb{C}^{d_0d \times d_0d}\), \(B \in \mathbb{C}^{d_0d \times d_0}\), \(C \in \mathbb{C}^{d_0 \times d_0d}\), and \(D \in \mathbb{R}^{d_0\times d_0}\) are shared across all dimensions of the input \(v_{w_t}\). To efficiently utilize the SSM in structured sequence modeling, a discrete-time formulation is required. The S4 model \citep{GuGR22} employs the bilinear transformation \citep{tustin1947method} to discretize the continuous-time system, leading to the following discrete-time state-space representation, which we refer to as an S4 cell:
\begin{align}
    \label{eq_S43}
    h_t &= \tilde{A} h_{t-1} + \tilde{B} v_{w_t},\\
    \label{eq_S44}
    y_t &= \tilde{C} h_t + \tilde{D} v_{w_t},
\end{align}
where the discretized parameters are given by:\\
\(\tilde{A} = (I - \Delta  / 2 \cdot A)^{-1}(I + \Delta  / 2 \cdot A)\in  \mathbb{C}^{d_0d \times d_0d}\), \(\tilde{B} = (I - \Delta  / 2 \cdot A)^{-1} \Delta  B \in  \mathbb{C}^{d_0d \times d_0}\), \(\tilde{C} = C \in  \mathbb{C}^{d_0 \times d_0d}\), \(\tilde{D} = D \in  \mathbb{R}^{d_0\times d_0}\), and \(\Delta \in \mathbb{R}^{d_0}\) is a fixed step size that represents the resolution of the input.
{Based on Eqs.\ (\ref{eq_S43}) and (\ref{eq_S44}), we can represent the S4 cell as an input--output mapping \(\mathcal{F}\), as in Eq.\ (\ref{eq:orig}), formally defined as:
\begin{equation}
\begin{split}
\mathcal{F} \colon \quad &\mathbb{R}^{d_0+d_0d} \;\to\; \mathbb{R}^{d_0}\\
&\hspace{-6mm}(v_{w_t}, h_{t-1}) \mapsto \tilde{C} \bigl(\tilde{A}h_{t-1} + \tilde{B}v_{w_t}\bigr)+ \tilde{D}v_{w_t},
\end{split}
\label{eqn:SSM}
\end{equation}
where the input is \(x = (v_{w_t}, h_{t-1})\in\mathbb{R}^{d_0+d_0d}\) and the output is \(y_t= \mathcal{F}(v_{w_t}, h_{t-1})=\tilde{C}\bigl(\tilde{A}h_{t-1} + \tilde{B}v_{w_t}\bigr) + \tilde{D}v_{w_t} \in \mathbb{R}^{d_0}\).
}\\
{In the following, we provide an explicit formula for the GBM of the S4 cell. A detailed proof of this result can be found in the Appendix \ref{sec:proofs}.}

\paragraph{Proposition 3:}

Consider the map $\mathcal{F}$ describing the input-output model of an S4 cell defined in Eq.\eqref{eqn:SSM}. The GBM of the map $\mathcal{F}$ can be expressed as follows: 
\[\text{For all }(i,j) \in  \{1,\ldots,d_0\} \times \{1,\ldots,d_0+d_0d\} \]
\begin{equation}\label{eq:GBMS4}
(\mathcal{M})_{i,j} = 
\begin{cases}
     \|\big(\tilde{C}.\tilde{B}\big)_{i,j} +\tilde{D}_{ij}\| \quad \text{if} \quad\hspace{1.5mm} 1\leq j \leq d_0\\
    \|\big(\tilde{C}.\tilde{A}\big)_{i,j-d_0}\| \hspace{2mm} \text{if} \hspace{2mm}  d_0<j \leq d_0+d_0d
\end{cases}
\end{equation}
\hspace{-4mm}\textbf{Convolutional Neural Network (CNN)}\\
TextCNN \citep{Kim14f} applies a one-dimensional convolutional operation followed by a max-pooling layer to extract meaningful representations.\\
{Given a sentence of \(N\) input word vectors \(v_x = \langle v_{w_1}, v_{w_2}, \ldots,v_{w_N} \rangle\), where each \(v_{w_t} \in \mathbb{R}^{d_0}\) denotes the embedding of the word at position \( t \).}\\
The convolutional layer and the max-pooling layer are defined as follows:\\
\(\text{For } k_i\in\mathcal{K}=\{k_1,\ldots,k_m\}\subset \mathbb{Z}_{\geq2}\) and \(t =  \{1,\ldots,N-k_i+1\} \)  
\begin{equation*}\label{eq:conv}
    \mathcal{C}^{(k_i)}(v_x)_t = \phi \left( b^{(k_i)}  + \sum_{l=0}^{k_i-1} W^{(k_i)}_{:,:,l}  v_{w_{t+l}} \right) , 
\end{equation*} 
\begin{equation}\label{eq:pool}
    \mathcal{P}^{(k_i)}(\mathcal{C}^{(k_i)}(v_x)) = \max_{t=1}^{N-k_i+1} \left( \mathcal{C}^{(k_i)}(v_x)_t \right),  
\end{equation}
where, $k_i$ is the kernel size, $\phi$ is the ReLU activation function,  $b^{(k_i)}\in \mathbb{R}^{d}$ is the bias, $W^{(k_i)}\in \mathbb{R}^{d \times d_0\times k_i}$ is weight matrix for the convolutional filter of size $k_i$, and $v_{w_{t+l}}$ is the word embedding at position $t+l$. Based on Eq.(\ref{eq:pool}), we can represent the CNN layer as an input-output mapping \(\mathcal{F}\), ad in Eq.\ (\ref{eq:orig}), formally defined as:
\begin{equation}\label{eq:cnn}
    \begin{array}{l}
    \mathcal{F}: \quad \mathbb{R}^{{N d_0}} \to \mathbb{R}^{|\mathcal{K}|.d} \\
     \quad \quad \quad v_{x}\quad \mapsto \bigoplus_{k=k_1}^{k_m}\mathcal{P}^{(k)}\circ\mathcal{C}^{(k)}(v_x)
    \end{array}
\end{equation}
where \( \bigoplus \) represents the concatenation operation, and \(|\mathcal{K}|\) is the cardinal of the set \(\mathcal{K}\).\\
The following result presents an explicit formula for the GBM of the CNN layer, with a detailed proof provided in Appendix \ref{sec:proofs}.   
\paragraph{Proposition 4:}

Consider the map \(\mathcal{F}\) defined in Eq.\eqref{eq:cnn}. The GBM of the map $\mathcal{F}$ can be expressed as follows:
\[\text{For all }(i,j) \in  \{1,\ldots,|\mathcal{K}|.d\} \times \{1,\ldots,Nd_0\} \] 
\begin{equation}\label{eq:GBMcnn}
    (\mathcal{M})_{i,j} = \max_{t=1}^{j} \|\,
   W^{\bigl(k_{\alpha(i,1,d)}\bigr)}_{\,\beta(i\,,\,1\,,\,d),\,\beta(j\,,\,t\,,\,d_0),\,\alpha(j\,,\,t\,,\,d_0)}
\| ,
\end{equation}
where, \[
\beta(i,a,d)= 1 + \bigl((i - a) \bmod d\bigr),\]
\[
\alpha(i,a,d) = \left\lfloor \frac{i - a}{d}\right\rfloor + 1,\]
\(\lfloor.\rfloor\) is the floor function, and the modulo operation $\bmod$ computes the remainder of the division of $(i-a)$ by $d$.




\begin{table*}[htbp]
    \centering
    \setlength{\tabcolsep}{5.2pt} 
    \begin{tabular}{llcccccccc}
        \toprule
         \multirow{2}{*}{Dataset} & \multirow{2}{*}{Defense} & \multicolumn{4}{c}{CNN} & \multicolumn{4}{c}{BiLSTM}\\
         \cmidrule(lr){3-6} \cmidrule(lr){8-10}
          ~ & ~ & Clean & PWWS & GA & PSO & Clean & PWWS & GA & PSO \\
         \midrule
         \multirow{7}{*}{\textit{IMDB}} &
         Standard & \underline{89.7} & ~~0.6 & ~~2.6 & ~~1.4 & \underline{89.1} & ~~0.2 & ~~1.6 &  ~~0.3 \\
         &IBP \citep{jia19certified}& 81.7 & \underline{75.9} & \underline{76.0} & \underline{75.9} & 77.6 & 67.5 & 67.8 & 67.6 \\
         & ATFL \citep{wang21adversarial}& 85.0 & 63.6 & 66.8 & 64.7 & 85.1 & 72.2 & 75.5 & 74.0\\
         & SEM \citep{wang21natural}& 87.6 & 62.2 & 63.5 & 61.5 & 86.8 & 61.9 & 63.7 & 62.2\\
         & ASCC \citep{dong21towards} & 84.8 & 74.0 & 75.5 & 74.5 & 84.3 & \underline{74.2} & \underline{76.8} & \underline{75.5}\\
         \midrule
         & GBM & \textbf{90.2} & \textbf{76.3} & \textbf{76.6} & \textbf{84.3} & \textbf{89.6} & \textbf{76.8} & \textbf{77.8} & \textbf{84.3} \\
         & & & \textcolor{blue}{\(\uparrow\)0.4} & \textcolor{blue}{\(\uparrow\)0.6} & \textcolor{blue}{\(\uparrow\)8.4} & & \textcolor{blue}{\(\uparrow\)2.6} & \textcolor{blue}{\(\uparrow\)1.0} & \textcolor{blue}{\(\uparrow\)8.8} \\
         \midrule
         \multirow{7}{*}{\textit{\shortstack{Yahoo!\\ Answers}}} &
         Standard & \underline{72.6} & ~~6.8 & ~~7.2 & ~~4.9 & \textbf{74.7} & 12.2 & ~~9.6 & ~~6.5 \\
         & IBP \citep{jia19certified} & 63.1 & 54.9 & 54.9 & 54.8 & 54.3 & 47.3 & 47.6 & 47.0 \\
         & ATFL \citep{wang21adversarial} & 72.5 & \underline{62.5} & \underline{63.1} & \underline{62.5} & \underline{73.6} & \underline{61.7} & 60.8 & 60.3\\ 
         & SEM \citep{wang21natural}& 70.1 & 53.8 & 52.4 & 51.9 & 72.3 & 57.0 & 56.1 & 55.4 \\
         & ASCC \citep{dong21towards} & 69.0 & 58.4 & 59.6 & 58.5 & 70.7 & \underline{61.7} & \underline{62.3} & \underline{61.9} \\
         \midrule
         & GBM & \textbf{72.8} & \textbf{66.2} & \textbf{66.1} & \textbf{67.3} & 73.0 & \textbf{67.0} & \textbf{67.0} & \textbf{68.6} \\
         &  & & \textcolor{blue}{\(\uparrow\)3.7} & \textcolor{blue}{\(\uparrow\)3.0} & \textcolor{blue}{\(\uparrow\)4.8} & & \textcolor{blue}{\(\uparrow\)5.3} & \textcolor{blue}{\(\uparrow\)4.7} & \textcolor{blue}{\(\uparrow\)6.7} \\
         \bottomrule
    \end{tabular}
    \caption{The clean accuracy (CA) (\%) for standard training and Accuracy Under Attack (AUA) (\%) against various adversarial attacks on two datasets for CNN and BiLSTM. The columns of \textit{Clean} denote the CA on the entire original testing set. The highest accuracy against the corresponding attack on each column is highlighted in \textbf{bold}, while the second one is highlighted in \underline{underline}. The last row of each block indicates the gains of the accuracy between our method and the best baseline.}
    \label{tab:defense_bilstm}
    \vspace{-.1in}
\end{table*}
\section{Experiments}

{This section provides an experimental evaluation of our proposed GBM approach, comparing its performance with four defense baselines against three distinct adversarial attacks on the BiLSTM and CNN models, and with eleven defense baselines against the TextFooler attack on the BERT model. We also study the robustness of the S4 model against three distinct adversarial attacks. {Additionally, we emphasize the scalability of the GBM approach in terms of time efficiency, and we provide visual evidence of its resilience to adversarial perturbations.}

\subsection{Experimental Setup}

\textbf{Datasets:} We evaluate our method on two benchmark datasets: \textit{IMDB} \citep{maas11learning}: A binary sentiment classification dataset with 25,000 movie reviews for training and 25,000 for testing. \textit{Yahoo! Answers} \citep{zhang15character}: A large-scale topic classification dataset with 1,400,000 training samples and 50,000 testing samples across 10 classes.

\paragraph{Attack Methods:} To evaluate defense efficacy, we employ five advanced adversarial attacks: GA (Genetic Attack) \citep{alzantot-etal-2018-generating}, PWWS (Probability Weighted Word Saliency) \citep{shuhuai19generating}, PSO (Particle Swarm Optimization) \citep{yuan20pso}, and TextFooler \citep{jin2020bert}. {PWWS uses synonyms from WordNet; TextFooler and GA use counter-fitted embeddings; PSO relies on sememe-based substitutions from HowNet.} Given the computational cost of textual adversarial attacks, we generate adversarial examples using 1000 randomly sampled test instances from each dataset. For fair comparison with Text-CRS \citep{zhang2024text}, we sample 500 instances from the IMDB test set and assess model robustness against the TextFooler attack. All attacks are implemented using the OpenAttack framework \citep{zeng-etal-2021-openattack}, and TextAttack framework \citep{morris2020textattack}.

\paragraph{Perturbation Setting:} In line with \citep{jia19certified} and \citep{dong21towards}, we set \(k=8\) as the number of nearest neighbors within a Euclidean distance of \({d_e}=0.5\) in the GloVe embedding space. Additionally, for BERT model, following \citep{zhang2024text}, we adopt the default parameters of \citep{jin2020bert} to ensure a fair comparison.
\paragraph{Model Setting:} We employ pre-trained GloVe embeddings to map words into a 300-dimensional vector space, serving as the embedding layer for Bi-LSTM, CNN, and S4 models. Additionally, we utilize the pre-trained 12-layer bert-base-uncased model, which generates 768-dimensional hidden representations per token, as the embedding layer for Bi-LSTM and CNN models. Further implementation details are provided in Appendix \ref{sec:Exp}. 
\paragraph{Evaluation Metrics:} We use the following metrics to evaluate our models: Clean Accuracy (CA) denotes model’s classification accuracy on the clean test dataset, and Accuracy Under Attack (AUA) is the model’s prediction accuracy under specific adversarial attack methods.

\subsection{Main Results}

\paragraph{Performance on CNN and BiLSTM:} 

Table \ref{tab:defense_bilstm} presents the classification performance of the BiLSTM and CNN models on the \textit{IMDB} and \textit{Yahoo! Answers} datasets, where each row corresponds to a defense method and each column to an attack method. A defense is considered more effective if it preserves higher AUA while minimizing performance degradation on clean samples compared to standard training.\\
Our proposed method consistently outperforms existing defenses, particularly against the PSO attack, achieving improvements of up to \textbf{8.8\%} on \textit{IMDB} and \textbf{6.7\%} on \textit{Yahoo! Answers} with the BiLSTM model, and up to \textbf{8.4\%} on \textit{IMDB} and \textbf{4.8\%} on \textit{Yahoo! Answers} with the CNN model. Furthermore, our approach attains the highest CA on both datasets with CNN and nearly the best CA on \textit{Yahoo! Answers} with the BiLSTM model. These results highlight its effectiveness in enhancing model robustness while maintaining strong generalization performance in both clean and adversarial settings.

\paragraph{Performance on S4:}
Figure \ref{fig:defense_s4} shows how the S4 model performs on the \textit{IMDB} dataset under normal conditions and different adversarial attacks. Since this is the first study evaluating S4’s robustness against such attacks, there are no previous baselines for direct comparison.\\
Our results reveal that the standard S4 model achieves the highest CA (\textbf{88.1\%}), but it struggles against adversarial attacks. By applying our GBM defense, we significantly improve S4’s resistance to attacks, achieving the highest AUA for PWWS (\textbf{49.4\%}), PSO (\textbf{42.6\%}), and TextFooler (\textbf{52.6\%}). These findings demonstrate that our method effectively strengthens S4’s defense against adversarial perturbations while maintaining high CA.
\begin{figure}[ht]
    \centering
    \includegraphics[width=\columnwidth]{ 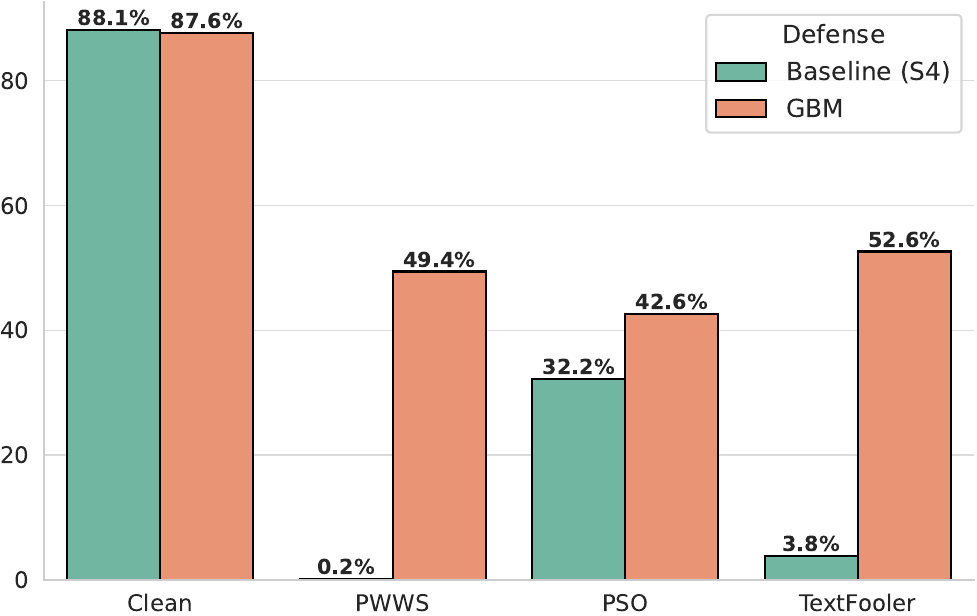}
    \caption{The CA (\%) for standard training and AUA (\%) against various adversarial attacks on IMDB dataset for S4 model.}
    \label{fig:defense_s4}
    \vspace{-.2in}
\end{figure}
\paragraph{Performance on BERT:}


{The results presented in Table \ref{tab:defense_bert} demonstrate that the GBM-based method achieves superior robustness against the TextFooler attack compared to baseline approaches across both the BiLSTM model (utilizing BERT embeddings) and the BERT model. Specifically, our approach outperforms Text-CRS, achieving improvements of \textbf{{6.2\%}} against TextFooler. These findings highlight the effectiveness of the proposed methodology in enhancing model resilience relative to existing techniques in the literature.} \\
\begin{table}[h]
    \vspace{-.2in}
    \centering
    \resizebox{\columnwidth}{!}{
    \begin{tabular}{lcc}
        \toprule
        {Defense} & {Clean} & {TextFooler}\\
        \midrule
        baseline (BERT) & 92.1 & 10.3 \\
        MixADA \citep{si2020better}  & 91.9 & 19.0  \\
        PGD-K \citep{madry18towards}  & \underline{93.2} & 26.0 \\
        FreeLB \citep{zhu2019freelb}  & 93.0 & 29.0  \\
        TA-VAT \citep{li2020tavattokenawarevirtualadversarial}  & 93.0 & 28.0 \\
        InfoBERT \citep{wang2020infobert} & 92.0 & 19.0\\
        DNE \citep{zhou2020defense} & 90.4 & 28.0 \\
        ASCC \citep{dong21towards} & 87.8 & 19.4 \\
        SAFER \citep{yeetal2020safer}  & \textbf{93.5} & 9.5 \\
        RanMASK \citep{zeng2021certified} & \underline{93.2} & 22.0 \\
        FreeLB++ \citep{li2021searching} & \underline{93.2} & 45.3 \\
        Text-CRS \citep{zhang2024text} & 91.5 & \underline{84.4} \\
        \bottomrule
        GBM & {92.1} & \textbf{90.6} \\
        & & \textcolor{blue}{\(\uparrow\)6.2}\\
        \bottomrule
    \end{tabular}}
    \caption{The CA (\%) for standard training and AUA (\%) against TextFooler attack on IMDB dataset for BERT models.}
    \label{tab:defense_bert}
    \vspace{-5mm}
\end{table}

\subsection{Time Efficiency of the GBM Approach}
Efficiency plays a critical role in evaluating defense mechanisms, particularly for large-scale datasets and complex models. Table \ref{tab:time_memory} presents the training time per epoch for both GBM and IBP \cite{jia19certified} defense methods on the IMDB dataset. Notably, the IBP defense method demands excessively more training time for BiLSTM models, primarily due to the heavy overhead imposed by certified constraints. In contrast, the comparison shows the significantly higher computational efficiency of the GBM approach, reducing training time by more than \textbf{{11 times}} compared to IBP.
\begin{table}[ht]
    \centering
    \resizebox{\columnwidth}{!}{
    \begin{tabular}{@{}lccc@{}}
        \toprule
        Dataset & Model & Defense & Time by epoch (min) \\
        \midrule
        \multirow{4}{*}{\textit{IMDB}} & \multirow{2}{*}{\textit{BiLSTM}} & IBP  & 04:53 \\
        & & GBM & \textbf{00:25} \\
        \cmidrule(lr){2-4}
         & \multirow{2}{*}{{\textit{CNN}}} & {IBP}  & {00:05} \\
        & & {GBM} & {\textbf{00:02}} \\
        \bottomrule
    \end{tabular}}
    \caption{Training time per epoch for IBP and GBM defense methods on the IMDB dataset.}
    \label{tab:time_memory}
    \vspace{-.2in}
\end{table}

\begin{figure*}[ht]
    \centering
    \includegraphics[width=11cm]{ 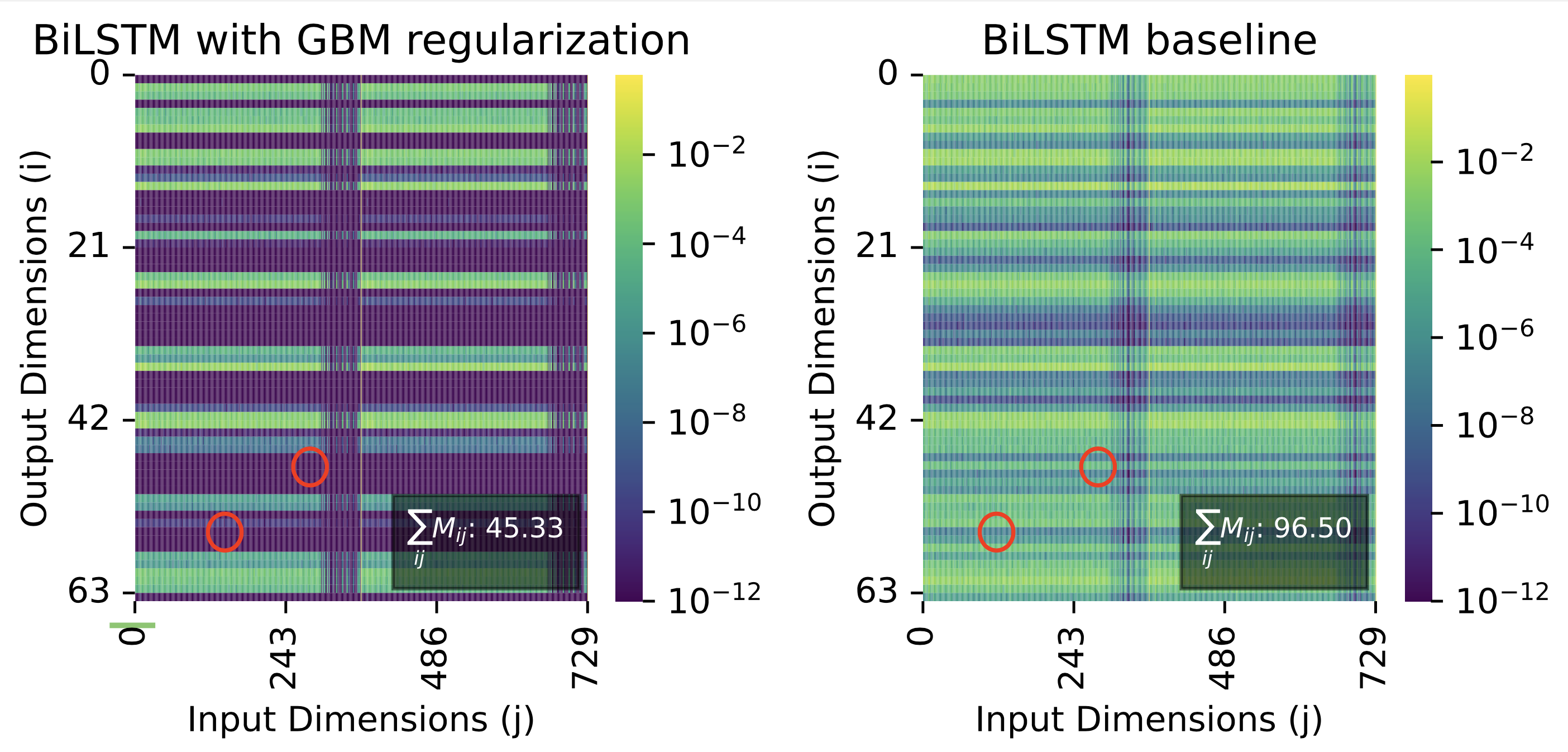}
    \caption{Comparison of GBM $\mathcal{M}$ in BiLSTM Models on the \textit{IMDB} dataset with and without GBM Regularization. The left heatmap corresponds to the model regularized with GBM, while the heatmap on the right represents the unregularized (baseline) model. The red circles highlight representative examples where individual entries of the GBM matrix $\mathcal{M}_{ij}$ are strongly minimized in the regularized model compared to the baseline.}
    \label{fig:heatmap}
\end{figure*}
\begin{figure*}[ht]
    \centering
    \includegraphics[width=14cm]{ 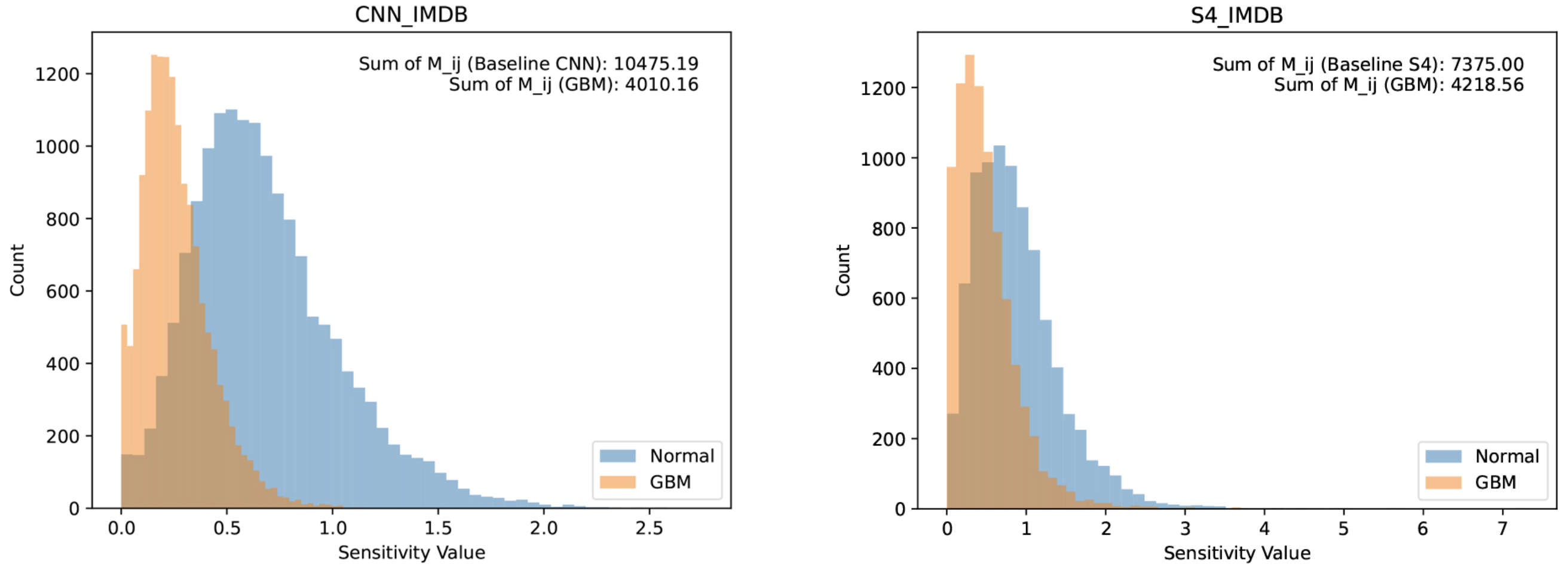}
    \caption{Comparison of the distribution of elements in the GBM $\mathcal{M}$ for CNN and S4 models on the \textit{IMDB} dataset.}
    \label{fig:dist}
\end{figure*}

\subsection{Efficiency of GBM against Adversarial perturbations}
\textbf{BiLSTM model:} 
Figure~\ref{fig:heatmap} shows sensitivity heatmaps for the final hidden state of a BiLSTM model trained on the \textit{IMDB} dataset. The heatmap on the left corresponds to the model trained with GBM regularization, while the right heatmap represents the baseline model without regularization. These heatmaps visualize how much each input dimension influences each output dimension, based on the GBM matrix $\mathcal{M}$. The color scale (logarithmic) highlights the magnitude of the components of $\mathcal{M}$, where brighter colors indicate stronger influence, i.e., larger values of a given component of $\mathcal{M}$. We observe that the total sensitivity, computed as $\sum_{i,j} \mathcal{M}_{ij}$, is significantly lower in the GBM-regularized model (45.33) compared to the baseline model (96.50). This suggests that GBM reduces the model’s reaction to small input changes, thereby promoting stability and improving robustness to perturbations.\\


\hspace{-4mm}\textbf{CNN and S4 models:} 
Figure~\ref{fig:dist} illustrates the distribution of the values of the GBM matrix $\mathcal{M}$ for CNN and S4 models trained on the \textit{IMDB} dataset, with and without GBM regularization. In both cases, the distributions for GBM-regularized models (orange) are clearly shifted toward lower values of $\mathcal{M}_{ij}$ compared to their baseline counterparts (blue), indicating a reduction in the overall magnitude of input-output interactions. The total sensitivity of the matrix, computed as the sum of all its components $\sum_{i,j} \mathcal{M}_{ij}$, also decreases significantly under GBM regularization. Specifically, for the CNN model, the total sum drops from 10475.19 to 4010.16, while for the S4 model, it decreases from 7375.00 to 4218.56. These results highlight the effectiveness of GBM to improve the robustness against adversarial inputs.

\section{Discussion}
{Our results demonstrate that minimizing the GBM yields two advantages : (1) enhanced robustness against adversarial attacks and (2) improved generalization on clean data. This is achieved by introducing a regularization term, which encourages the model to learn smoother decision boundaries. Adversarial methods (e.g., GA, PWWS, PSO, TextFooler) exploit small input perturbations (e.g., synonym substitutions) to provoke large output deviations. By reducing sensitivity (via low GBM), gradient-based or search-driven attack strategies face significant difficulty in identifying effective perturbations under practical modification limits. Crucially, a controlled GBM penalty preserves the model’s ability to learn discriminative features, retaining high accuracy on clean data. Thus, reducing sensitivity by minimizing the GBM ensures a crucial balance between adversarial robustness and performance on clean inputs. Further analysis can be found in the Appendix \ref{sec:Exp}.}
\section{Conclusion}
In this work, we introduce a novel method, Growth Bound Matrices (GBM), which provides tight bounds on the input-output variations of diverse models. By minimizing the change in a model's predictions when substituting a word with one of its synonyms, GBM ensures robustness against such perturbations. Empirical evaluations demonstrate the effectiveness of our approach, revealing significant improvements in model robustness.\\
For future work, we plan to investigate the applicability of GBM across additional domains, including speech recognition \citep{sajjad2020clustering} and control systems \citep{jouret2023safety}. Furthermore, we aim to explore GBM's potential for State Space Model S6 (Mamba) \citep{gu2023mamba}, thereby extending its contribution to robust model design.

\section*{Limitations}
In this work, we focus on sequence-level classification, where the objective is to determine the overall class of an entire text sequence. Our approach is not directly applicable to token-level classification tasks, which require predicting the class of each individual word within a sequence. However, it is well-suited for word substitution attacks with fixed-length perturbations. Future work should investigate extending our method to enhance robustness against a broader range of word-level adversarial attacks, including variable-length perturbations such as word deletion or removal, thereby improving its applicability to more fine-grained text classification tasks.
\bibliography{acl_latex.bbl}

\begin{thebibliography}{51}
\providecommand{\natexlab}[1]{#1}

\bibitem[{Alzantot et~al.(2018)Alzantot, Sharma, Elgohary, Ho, Srivastava, and Chang}]{alzantot-etal-2018-generating}
Moustafa Alzantot, Yash Sharma, Ahmed Elgohary, Bo-Jhang Ho, Mani Srivastava, and Kai-Wei Chang. 2018.
\newblock Generating natural language adversarial examples.
\newblock In \emph{Proceedings of the 2018 Conference on Empirical Methods in Natural Language Processing}.

\bibitem[{Dong et~al.(2021)Dong, Luu, Ji, and Liu}]{dong21towards}
Xinshuai Dong, Anh~Tuan Luu, Rongrong Ji, and Hong Liu. 2021.
\newblock Towards robustness against natural language word substitutions.
\newblock In \emph{9th International Conference on Learning Representations}.

\bibitem[{Ebrahimi et~al.(2018)Ebrahimi, Rao, Lowd, and Dou}]{javid18hotflip}
Javid Ebrahimi, Anyi Rao, Daniel Lowd, and Dejing Dou. 2018.
\newblock {H}ot{F}lip: White-box adversarial examples for text classification.
\newblock In \emph{Proceedings of the 56th Annual Meeting of the Association for Computational Linguistics}.

\bibitem[{Eger and Benz(2020)}]{eger2020hero}
Steffen Eger and Yannik Benz. 2020.
\newblock From hero to z{\'e}roe: A benchmark of low-level adversarial attacks.
\newblock In \emph{Proceedings of the 1st conference of the Asia-Pacific chapter of the association for computational linguistics and the 10th international joint conference on natural language processing}, pages 786--803.

\bibitem[{Ge et~al.(2019)Ge, Zhang, Wei, and Zhou}]{ge-etal-2019-automatic}
Tao Ge, Xingxing Zhang, Furu Wei, and Ming Zhou. 2019.
\newblock \href {https://doi.org/10.18653/v1/P19-1609} {Automatic grammatical error correction for sequence-to-sequence text generation: An empirical study}.
\newblock In \emph{Proceedings of the 57th Annual Meeting of the Association for Computational Linguistics}, pages 6059--6064, Florence, Italy. Association for Computational Linguistics.

\bibitem[{Goodfellow et~al.(2015)Goodfellow, Shlens, and Szegedy}]{goodfellow2014explaining}
Ian~J. Goodfellow, Jonathon Shlens, and Christian Szegedy. 2015.
\newblock Explaining and harnessing adversarial examples.
\newblock In \emph{3rd International Conference on Learning Representations}.

\bibitem[{Gu and Dao(2023)}]{gu2023mamba}
Albert Gu and Tri Dao. 2023.
\newblock Mamba: Linear-time sequence modeling with selective state spaces.
\newblock \emph{arXiv preprint arXiv:2312.00752}.

\bibitem[{Gu et~al.(2022)Gu, Goel, and R{\'{e}}}]{GuGR22}
Albert Gu, Karan Goel, and Christopher R{\'{e}}. 2022.
\newblock Efficiently modeling long sequences with structured state spaces.
\newblock In \emph{ICLR}.

\bibitem[{Hochreiter and Schmidhuber(1997)}]{hochreiter1997long}
Sepp Hochreiter and J{\"u}rgen Schmidhuber. 1997.
\newblock Long short-term memory.
\newblock \emph{Neural computation}, 9(8):1735--1780.

\bibitem[{Huang et~al.(2019)Huang, Stanforth, Welbl, Dyer, Yogatama, Gowal, Dvijotham, and Kohli}]{huang19achieving}
Po-Sen Huang, Robert Stanforth, Johannes Welbl, Chris Dyer, Dani Yogatama, Sven Gowal, Krishnamurthy Dvijotham, and Pushmeet Kohli. 2019.
\newblock Achieving verified robustness to symbol substitutions via interval bound propagation.
\newblock In \emph{Proceedings of the 2019 Conference on Empirical Methods in Natural Language Processing and the 9th International Joint Conference on Natural Language Processing}.

\bibitem[{Ivgi and Berant(2021)}]{ivgi21achieving}
Maor Ivgi and Jonathan Berant. 2021.
\newblock Achieving model robustness through discrete adversarial training.
\newblock In \emph{Proceedings of the 2021 Conference on Empirical Methods in Natural Language Processing}, pages 1529--1544.

\bibitem[{Jia et~al.(2019)Jia, Raghunathan, G{\"{o}}ksel, and Liang}]{jia19certified}
Robin Jia, Aditi Raghunathan, Kerem G{\"{o}}ksel, and Percy Liang. 2019.
\newblock Certified robustness to adversarial word substitutions.
\newblock In \emph{Proceedings of the 2019 Conference on Empirical Methods in Natural Language Processing and the 9th International Joint Conference on Natural Language Processing}, pages 4127--4140.

\bibitem[{Jin et~al.(2020)Jin, Jin, Zhou, and Szolovits}]{jin2020bert}
Di~Jin, Zhijing Jin, Joey~Tianyi Zhou, and Peter Szolovits. 2020.
\newblock Is bert really robust? a strong baseline for natural language attack on text classification and entailment.
\newblock In \emph{Proceedings of the AAAI conference on artificial intelligence}, volume~34, pages 8018--8025.

\bibitem[{Jouret et~al.(2023)Jouret, Saoud, and Olaru}]{jouret2023safety}
Louis Jouret, Adnane Saoud, and Sorin Olaru. 2023.
\newblock Safety verification of neural-network-based controllers: a set invariance approach.
\newblock \emph{IEEE Control Systems Letters}, 7:3842--3847.

\bibitem[{Kim(2014)}]{Kim14f}
Yoon Kim. 2014.
\newblock \href {https://arxiv.org/abs/1408.5882} {Convolutional neural networks for sentence classification}.
\newblock \emph{CoRR}, abs/1408.5882.

\bibitem[{Li and Qiu(2020{\natexlab{a}})}]{li2020tavat}
Linyang Li and Xipeng Qiu. 2020{\natexlab{a}}.
\newblock Tavat: Token-aware virtual adversarial training for language understanding.
\newblock \emph{arXiv preprint arXiv:2004.14543}.

\bibitem[{Li and Qiu(2020{\natexlab{b}})}]{li2020tavattokenawarevirtualadversarial}
Linyang Li and Xipeng Qiu. 2020{\natexlab{b}}.
\newblock \href {https://arxiv.org/abs/2004.14543} {Tavat: Token-aware virtual adversarial training for language understanding}.
\newblock \emph{Preprint}, arXiv:2004.14543.

\bibitem[{Li et~al.(2021)Li, Xu, Zeng, Li, Zheng, Zhang, Chang, and Hsieh}]{li2021searching}
Zongyi Li, Jianhan Xu, Jiehang Zeng, Linyang Li, Xiaoqing Zheng, Qi~Zhang, Kai-Wei Chang, and Cho-Jui Hsieh. 2021.
\newblock Searching for an effective defender: Benchmarking defense against adversarial word substitution.
\newblock \emph{arXiv preprint arXiv:2108.12777}.

\bibitem[{Liu et~al.(2021)Liu, Arnon, Lazarus, Strong, Barrett, Kochenderfer et~al.}]{liu2021algorithms}
Changliu Liu, Tomer Arnon, Christopher Lazarus, Christopher Strong, Clark Barrett, Mykel~J Kochenderfer, et~al. 2021.
\newblock Algorithms for verifying deep neural networks.
\newblock \emph{Foundations and Trends{\textregistered} in Optimization}, 4(3-4):244--404.

\bibitem[{Maas et~al.(2011)Maas, Daly, Pham, Huang, Ng, and Potts}]{maas11learning}
Andrew~L. Maas, Raymond~E. Daly, Peter~T. Pham, Dan Huang, Andrew~Y. Ng, and Christopher Potts. 2011.
\newblock Learning word vectors for sentiment analysis.
\newblock In \emph{The 49th Annual Meeting of the Association for Computational Linguistics: Human Language Technologies}.

\bibitem[{Madry et~al.(2018)Madry, Makelov, Schmidt, Tsipras, and Vladu}]{madry18towards}
Aleksander Madry, Aleksandar Makelov, Ludwig Schmidt, Dimitris Tsipras, and Adrian Vladu. 2018.
\newblock Towards deep learning models resistant to adversarial attacks.
\newblock In \emph{6th International Conference on Learning Representations}.

\bibitem[{Maheshwary et~al.(2021)Maheshwary, Maheshwary, and Pudi}]{rishabh21generating}
Rishabh Maheshwary, Saket Maheshwary, and Vikram Pudi. 2021.
\newblock Generating natural language attacks in a hard label black box setting.
\newblock In \emph{Thirty-Fifth {AAAI} Conference on Artificial Intelligence}.

\bibitem[{Meyer et~al.(2021)Meyer, Devonport, and Arcak}]{meyer2021interval}
Pierre-Jean Meyer, Alex Devonport, and Murat Arcak. 2021.
\newblock \emph{Interval reachability analysis: Bounding trajectories of uncertain systems with boxes for control and verification}.
\newblock Springer Nature.

\bibitem[{Morris et~al.(2020)Morris, Lifland, Yoo, Grigsby, Jin, and Qi}]{morris2020textattack}
John Morris, Eli Lifland, Jin~Yong Yoo, Jake Grigsby, Di~Jin, and Yanjun Qi. 2020.
\newblock Textattack: A framework for adversarial attacks, data augmentation, and adversarial training in nlp.
\newblock In \emph{Proceedings of the 2020 Conference on Empirical Methods in Natural Language Processing: System Demonstrations}, pages 119--126.

\bibitem[{Pei and Yue(2022)}]{pei2022generating}
Weiping Pei and Chuan Yue. 2022.
\newblock Generating content-preserving and semantics-flipping adversarial text.
\newblock In \emph{Proceedings of the 2022 ACM on Asia Conference on Computer and Communications Security}, pages 975--989.

\bibitem[{Pruthi et~al.(2019{\natexlab{a}})Pruthi, Dhingra, and Lipton}]{danish19combating}
Danish Pruthi, Bhuwan Dhingra, and Zachary~C. Lipton. 2019{\natexlab{a}}.
\newblock Combating adversarial misspellings with robust word recognition.
\newblock In \emph{Proceedings of the 57th Conference of the Association for Computational Linguistics}, pages 5582--5591.

\bibitem[{Pruthi et~al.(2019{\natexlab{b}})Pruthi, Dhingra, and Lipton}]{pruthi-etal-2019-combating}
Danish Pruthi, Bhuwan Dhingra, and Zachary~C. Lipton. 2019{\natexlab{b}}.
\newblock \href {https://doi.org/10.18653/v1/P19-1561} {Combating adversarial misspellings with robust word recognition}.
\newblock In \emph{Proceedings of the 57th Annual Meeting of the Association for Computational Linguistics}, pages 5582--5591, Florence, Italy. Association for Computational Linguistics.

\bibitem[{Qi et~al.(2024)Qi, Luo, Gao, Li, Tian, Ma, and Zhou}]{qi2024exploring}
Biqing Qi, Yang Luo, Junqi Gao, Pengfei Li, Kai Tian, Zhiyuan Ma, and Bowen Zhou. 2024.
\newblock Exploring adversarial robustness of deep state space models.
\newblock \emph{arXiv preprint arXiv:2406.05532}.

\bibitem[{Ren et~al.(2019)Ren, Deng, He, and Che}]{shuhuai19generating}
Shuhuai Ren, Yihe Deng, Kun He, and Wanxiang Che. 2019.
\newblock Generating natural language adversarial examples through probability weighted word saliency.
\newblock In \emph{Proceedings of the 57th Conference of the Association for Computational Linguistics}.

\bibitem[{Sajjad et~al.(2020)Sajjad, Kwon et~al.}]{sajjad2020clustering}
Muhammad Sajjad, Soonil Kwon, et~al. 2020.
\newblock Clustering-based speech emotion recognition by incorporating learned features and deep bilstm.
\newblock \emph{IEEE access}, 8:79861--79875.

\bibitem[{Shayegani et~al.(2023)Shayegani, Mamun, Fu, Zaree, Dong, and Abu-Ghazaleh}]{shayegani2023survey}
Erfan Shayegani, Md~Abdullah~Al Mamun, Yu~Fu, Pedram Zaree, Yue Dong, and Nael Abu-Ghazaleh. 2023.
\newblock Survey of vulnerabilities in large language models revealed by adversarial attacks.
\newblock \emph{arXiv preprint arXiv:2310.10844}.

\bibitem[{Si et~al.(2020)Si, Zhang, Qi, Liu, Wang, Liu, and Sun}]{si2020better}
Chenglei Si, Zhengyan Zhang, Fanchao Qi, Zhiyuan Liu, Yasheng Wang, Qun Liu, and Maosong Sun. 2020.
\newblock Better robustness by more coverage: Adversarial training with mixup augmentation for robust fine-tuning.
\newblock \emph{arXiv preprint arXiv:2012.15699}.

\bibitem[{Song et~al.(2021)Song, Yu, Peng, and Narasimhan}]{song2021universal}
Liwei Song, Xinwei Yu, Hsuan-Tung Peng, and Karthik Narasimhan. 2021.
\newblock Universal adversarial attacks with natural triggers for text classification.
\newblock In \emph{Proceedings of the 2021 Conference of the North American Chapter of the Association for Computational Linguistics: Human Language Technologies}, pages 3724--3733.

\bibitem[{Tustin(1947)}]{tustin1947method}
Arnold Tustin. 1947.
\newblock A method of analysing the behaviour of linear systems in terms of time series.
\newblock \emph{Journal of the Institution of Electrical Engineers-Part IIA: Automatic Regulators and Servo Mechanisms}, 94(1):130--142.

\bibitem[{Wang et~al.(2020{\natexlab{a}})Wang, Pei, Pan, Chen, Wang, and Li}]{wang2019t3}
Boxin Wang, Hengzhi Pei, Boyuan Pan, Qian Chen, Shuohang Wang, and Bo~Li. 2020{\natexlab{a}}.
\newblock T3: Tree-autoencoder constrained adversarial text generation for targeted attack.
\newblock In \emph{Conference on Empirical Methods in Natural Language Processing}.

\bibitem[{Wang et~al.(2020{\natexlab{b}})Wang, Wang, Cheng, Gan, Jia, Li, and Liu}]{wang2020infobert}
Boxin Wang, Shuohang Wang, Yu~Cheng, Zhe Gan, Ruoxi Jia, Bo~Li, and Jingjing Liu. 2020{\natexlab{b}}.
\newblock Infobert: Improving robustness of language models from an information theoretic perspective.
\newblock \emph{9th International Conference on Learning Representations (ICLR)}.

\bibitem[{Wang et~al.(2021{\natexlab{a}})Wang, Tang, Lou, and Xiong}]{wang21certified}
Wenjie Wang, Pengfei Tang, Jian Lou, and Li~Xiong. 2021{\natexlab{a}}.
\newblock Certified robustness to word substitution attack with differential privacy.
\newblock In \emph{Proceedings of the 2021 Conference of the North American Chapter of the Association for Computational Linguistics: Human Language Technologies}.

\bibitem[{Wang et~al.(2021{\natexlab{b}})Wang, Jin, Yang, and He}]{wang21natural}
Xiaosen Wang, Hao Jin, Yichen Yang, and Kun He. 2021{\natexlab{b}}.
\newblock Natural language adversarial defense through synonym encoding.
\newblock In \emph{Proceedings of the 37th Conference on Uncertainty in Artificial Intelligence}.

\bibitem[{Wang et~al.(2021{\natexlab{c}})Wang, Yang, Deng, and He}]{wang21adversarial}
Xiaosen Wang, Yichen Yang, Yihe Deng, and Kun He. 2021{\natexlab{c}}.
\newblock Adversarial training with fast gradient projection method against synonym substitution based text attacks.
\newblock In \emph{AAAI Conference on Artificial Intelligence}.

\bibitem[{Wang et~al.(2024)Wang, Wang, Chen, Wang, and Nguyen}]{wang2024generating}
Zimu Wang, Wei Wang, Qi~Chen, Qiufeng Wang, and Anh Nguyen. 2024.
\newblock Generating valid and natural adversarial examples with large language models.
\newblock In \emph{2024 27th International Conference on Computer Supported Cooperative Work in Design (CSCWD)}, pages 1716--1721. IEEE.

\bibitem[{Yang et~al.(2024)Yang, Meng, Zheng, and Wattenhofer}]{yang2024assessing}
Zeyu Yang, Zhao Meng, Xiaochen Zheng, and Roger Wattenhofer. 2024.
\newblock Assessing adversarial robustness of large language models: An empirical study.
\newblock \emph{arXiv preprint arXiv:2405.02764}.

\bibitem[{Ye et~al.(2020)Ye, Gong, and Liu}]{yeetal2020safer}
Mao Ye, Chengyue Gong, and Qiang Liu. 2020.
\newblock \href {https://doi.org/10.18653/v1/2020.acl-main.317} {{SAFER}: A structure-free approach for certified robustness to adversarial word substitutions}.
\newblock In \emph{Proceedings of the 58th Annual Meeting of the Association for Computational Linguistics}, pages 3465--3475, Online. Association for Computational Linguistics.

\bibitem[{Zang et~al.(2020)Zang, Qi, Yang, Liu, Zhang, Liu, and Sun}]{yuan20pso}
Yuan Zang, Fanchao Qi, Chenghao Yang, Zhiyuan Liu, Meng Zhang, Qun Liu, and Maosong Sun. 2020.
\newblock Word-level textual adversarial attacking as combinatorial optimization.
\newblock In \emph{Proceedings of the 58th Annual Meeting of the Association for Computational Linguistics}.

\bibitem[{Zeng et~al.(2021{\natexlab{a}})Zeng, Qi, Zhou, Zhang, Ma, Hou, Zang, Liu, and Sun}]{zeng-etal-2021-openattack}
Guoyang Zeng, Fanchao Qi, Qianrui Zhou, Tingji Zhang, Zixian Ma, Bairu Hou, Yuan Zang, Zhiyuan Liu, and Maosong Sun. 2021{\natexlab{a}}.
\newblock \href {https://doi.org/10.18653/v1/2021.acl-demo.43} {{O}pen{A}ttack: An open-source textual adversarial attack toolkit}.
\newblock In \emph{Proceedings of the 59th Annual Meeting of the Association for Computational Linguistics and the 11th International Joint Conference on Natural Language Processing: System Demonstrations}, pages 363--371, Online. Association for Computational Linguistics.

\bibitem[{Zeng et~al.(2021{\natexlab{b}})Zeng, Zheng, Xu, Li, Yuan, and Huang}]{zeng21certified}
Jiehang Zeng, Xiaoqing Zheng, Jianhan Xu, Linyang Li, Liping Yuan, and Xuanjing Huang. 2021{\natexlab{b}}.
\newblock Certified robustness to text adversarial attacks by randomized {[MASK]}.
\newblock In \emph{Findings of Association for Computational Linguistics}.

\bibitem[{Zeng et~al.(2021{\natexlab{c}})Zeng, Zheng, Xu, Li, Yuan, and Huang}]{zeng2021certified}
Jiehang Zeng, Xiaoqing Zheng, Jianhan Xu, Linyang Li, Liping Yuan, and Xuanjing Huang. 2021{\natexlab{c}}.
\newblock Certified robustness to text adversarial attacks by randomized [{MASK}].
\newblock \emph{arXiv preprint arXiv:2105.03743}.

\bibitem[{Zhang et~al.(2015)Zhang, Zhao, and LeCun}]{zhang15character}
Xiang Zhang, Junbo~Jake Zhao, and Yann LeCun. 2015.
\newblock Character-level convolutional networks for text classification.
\newblock In \emph{Advances in Neural Information Processing Systems}.

\bibitem[{Zhang et~al.(2024)Zhang, Hong, Hong, Huang, Wang, Ba, and Ren}]{zhang2024text}
Xinyu Zhang, Hanbin Hong, Yuan Hong, Peng Huang, Binghui Wang, Zhongjie Ba, and Kui Ren. 2024.
\newblock Text-crs: A generalized certified robustness framework against textual adversarial attacks.
\newblock In \emph{2024 IEEE Symposium on Security and Privacy (SP)}, pages 2920--2938. IEEE.

\bibitem[{Zhang et~al.(2021)Zhang, Albarghouthi, and D{'}Antoni}]{zhang-etal-2021-certified}
Yuhao Zhang, Aws Albarghouthi, and Loris D{'}Antoni. 2021.
\newblock \href {https://doi.org/10.18653/v1/2021.emnlp-main.82} {Certified robustness to programmable transformations in {LSTM}s}.
\newblock In \emph{Proceedings of the 2021 Conference on Empirical Methods in Natural Language Processing}, pages 1068--1083, Online and Punta Cana, Dominican Republic. Association for Computational Linguistics.

\bibitem[{Zhou et~al.(2020)Zhou, Zheng, Hsieh, Chang, and Huang}]{zhou2020defense}
Yi~Zhou, Xiaoqing Zheng, Cho-Jui Hsieh, Kai-wei Chang, and Xuanjing Huang. 2020.
\newblock Defense against adversarial attacks in nlp via dirichlet neighborhood ensemble.
\newblock \emph{arXiv preprint arXiv:2006.11627}.

\bibitem[{Zhu et~al.(2020)Zhu, Cheng, Gan, Sun, Goldstein, and Liu}]{zhu2019freelb}
Chen Zhu, Yu~Cheng, Zhe Gan, Siqi Sun, Tom Goldstein, and Jingjing Liu. 2020.
\newblock \href {https://openreview.net/forum?id=BygzbyHFvB} {Freelb: Enhanced adversarial training for natural language understanding}.
\newblock In \emph{8th International Conference on Learning Representations, {ICLR} 2020, Addis Ababa, Ethiopia, April 26-30, 2020}.

\end{thebibliography}
\newpage
\appendix
\section{Robustness via GBM}
\label{sec:RGBM}
\subsection{Proof of Proposition 1}

Consider the map \( \mathcal{F}: \mathbf{X} \rightarrow \mathbf{Y}\), where \(\mathbf{X} \subseteq \mathbb{R}^{n_x}\) and \(\mathbf{Y} \subseteq \mathbb{R}^{n_y}\). Each component \( \mathcal{F}^i: \mathbf{X}\subseteq \mathbb{R}^{n_x} \rightarrow \mathbb{R}\) is a scalar function. Given an input \( x \in \mathbf{X} \), a perturbation vector \( {\delta} \in \mathbb{R}^{n_x} \) and consider the perturbed input $x'\in \mathbf{X}$ such that $x'=x+{\delta}$. The mean value theorem guarantees that:
\[
\mathcal{F}^i(x) - \mathcal{F}^i(x') = \nabla\mathcal{F}^i(c)^T (x-x')
\]
for some point $c \in \mathbf{X}$ on the line segment between \(x\) to \(x'\). By taking norms on both sides, one gets:
\begin{eqnarray}
\label{eqn:prop1}
    \| \mathcal{F}^i(x) - \mathcal{F}^i(x') \|= \|\nabla\mathcal{F}^i(c)^T .(x-x')\| 
    \end{eqnarray}
Since \(\nabla\mathcal{F}^i(c) = \left( \dfrac{\partial \mathcal{F}^i}{\partial x^1}(c),\dots,\dfrac{\partial \mathcal{F}^i}{\partial x^{n_x}}(c)\right)^T\). By the definition of GBM (\ref{eq:GBM}), we have for \(c \in X\) that:
\[
\left\lVert \dfrac{\partial \mathcal{F}^i}{\partial x^j}(c)\right\rVert \leq (\mathcal{M})_{i,j} \quad \forall (i,j) \in \mathcal{I},
\]
where \( \mathcal{I} = \{1,\ldots,n_y\} \times \{1,\ldots,n_x\} \). Hence, one gets from Eq.(\ref{eqn:prop1}) that:
\begin{align*}
 \| \mathcal{F}^i(x) - \mathcal{F}^i(x') \| &=  \|\nabla\mathcal{F}^i(c)^T {\delta}\| \\ &\leq \sum\limits_{j=1}^{n_x}(\mathcal{M})_{i,j}\|{\delta_j}\|   
\end{align*}
where ${\delta_j}$ is the j-th component of the vector ${\delta}$. Then, the bound on the output of the perturbed input $x'$ can described as:
\begin{equation*}\label{eq:GBM_rows_eps}
    \centering
    \begin{aligned}
         \mathcal{F}^i(x) - \sum\limits_{j=1}^{n_x}(\mathcal{M})_{i,j}\|{\delta_j}\| &\leq \mathcal{F}^i(x') \\ &\hspace{-2.5mm}\leq
         \mathcal{F}^i(x) + \sum\limits_{j=1}^{n_x}(\mathcal{M})_{i,j}\|{\delta_j}\| 
    \end{aligned}
\end{equation*}
which concludes the proof. \hfill $\square$
\section{Relationship Between GBM and the Lipschitz Constant}

The Lipschitz constant corresponds to the largest value in the GBM, representing the maximum possible variation of outputs with respect to changes in inputs. The benefit of using the GBM over the Lipschitz constant is that while the Lipschitz constant provides a global upper bound on how much a function can stretch distances, regardless of direction, the GBM captures the variation in different directions, allowing tighter and more accurate bounds. For instance, the function might be more sensitive to changes in some directions than others, which a Lipschitz constant cannot express.

\paragraph{Proposition:} Consider the mapping \( \mathcal{F}: \mathbf{X} \rightarrow \mathbf{Y}  \)  and let a matrix \( \mathcal{M} \in \mathbb{R}^{n_y \times n_x} \) be its GBM. Then $L=\max\limits_i \max\limits_j (\mathcal{M})_{ij} $ is a Lipschitz constant of the map $\mathcal{F}$.

\paragraph{Proof:} Consider the map \( \mathcal{F}: \mathbf{X} \rightarrow \mathbf{Y}\), where \(\mathbf{X} \subseteq \mathbb{R}^{n_x}\) and \(\mathbf{Y} \subseteq \mathbb{R}^{n_y}\). Each component \( \mathcal{F}^i: \mathbf{X}\subseteq \mathbb{R}^{n_x} \rightarrow \mathbb{R}\) is a scalar function. Given two inputs \( x,x' \in \mathbf{X} \). 

If we consider the infinity norm, we have that $ \| \mathcal{F}(x) - \mathcal{F}(x') \|=\max\limits_i \| \mathcal{F}^i(x) - \mathcal{F}^i(x') \|$.
Now for each $i$, the mean value theorem guarantees that $$\| \mathcal{F}^i(x) - \mathcal{F}^i(x') \|= \|\nabla\mathcal{F}^i(c^i)^T\| \|(x-x')\|$$
for some point $c^i \in \mathbf{X}$ on the line segment between \(x\) to \(x'\). By taking norms on both sides, one gets:
\begin{align*}
    \| \mathcal{F}(x) - \mathcal{F}(x') \|&=\max\limits_i\| \mathcal{F}^i(x) - \mathcal{F}^i(x') \|\\&\leq \max\limits_i \|\nabla\mathcal{F}^i(c^i)^T\| \|x-x'\|  
    \end{align*}
Since \(\nabla\mathcal{F}^i(c^i) = \left( \dfrac{\partial \mathcal{F}^i}{\partial x^1}(c^i),\dots,\dfrac{\partial \mathcal{F}^i}{\partial x^{n_x}}(c^i)\right)^T\). We have that
\begin{align*}
    \| \mathcal{F}(x) - \mathcal{F}(x') \| &\leq\max\limits_i \|\nabla\mathcal{F}^i(c^i)^T\| \|x-x'\|\\& \leq \max\limits_i \max\limits_j \left\lVert \dfrac{\partial \mathcal{F}^i(c^i)}{\partial x^j} \right\lVert \|x-x'\|\\ & \leq \max\limits_i \max\limits_j (\mathcal{M})_{ij} \|x-x'\|\\&\leq L\|x-x'\|
    \end{align*}
Hence, $L=\max\limits_i \max\limits_j(\mathcal{M})_{ij} \|$ is a Lipschitz constant of the map $\mathcal{F}$.\hfill $\square$

\section{Auxiliary results}
\label{sec:AR}
In order to provide bounds on the sigmoid and hyperbolic tangent derivatives, we define the following result:

\paragraph{Proposition 5:} Consider an interval \([\underline{a}, \bar{a}]\subseteq \mathbb{R}\). For \(\varphi = \tanh'\text{ or } \sigma'\), the following holds:
\begin{align*} 
    \min_{a \in [\underline{a}, \bar{a}]} \varphi(a) &= 
        \min (\varphi(\underline{a}), \varphi(\bar{a})) \\ 
    \max_{a \in [\underline{a}, \bar{a}]} \varphi(a) &= \begin{cases}
        \varphi(0), & \text{if } 0 \in [\underline{a}, \bar{a}], \\
        \max (\varphi(\underline{a}), \varphi(\bar{a})), & \text{otherwise}.
    \end{cases}
\end{align*}

\paragraph{Proof of Proposition 5:}
Consider the scalar function $\varphi: \mathbb{R} \rightarrow \mathbb{R}$, with $\varphi=\tanh'$ or $\varphi=\sigma'$. To show the result, we analyse the variations of the function $\varphi$ on $\mathbb{R}$. Indeed, we have that $\varphi$ is:
\begin{itemize}
    \item non-decreasing on $(-\infty, 0]$ until reaching its global maxi$\operatorname{mum} \max _{x \in \mathbb{R}_{\infty}} \varphi(x)=\varphi(0)$;
    \item non-increasing on $[0,+\infty)$.
\end{itemize} The result in Proposition 5 follows then immediately.\hfill $\square$\\

The following result is adapted from \cite{liu2021algorithms}.
\paragraph{Proposition 6:} For $i\in \{1,\ldots,d\}$, consider the map $(v_{w_t},h_{t-1}) \mapsto T^i(v_{w_t},h_{t-1})$ described by 
\begin{small}
    \begin{equation*}
    T^i_{\gate} = \sum_{p=1}^{d_0} (\Theta^{(\gate)})_{pi}.v_{w_t}^{p} + \sum_{q=1}^{d} (U^{(\gate)})_{qi}.h_{t-1}^{q} + b^{(\gate)}_i
\end{equation*}
\end{small}
with $\Theta^{(gate)} \in \mathbb{R}^{d\times d_0} , U^{(gate)} \in \mathbb{R}^{d\times d}$ and $b^{(gate)} \in \mathbb{R}^{d}$ for $\gate \in \{f,I,g,o\}$. If $v_{w_t} \in [\underline{v},\overline{v}] \subseteq \mathbb{R}^{d_0}$ and $h_t \in [\underline{h}, \overline{h}] \subseteq \mathbb{R}^d$, then $T^i_{\gate}\in [\underline{T}^i,\overline{T}^i]$, where $\underline{T}^i$ and $\overline{T}^i$ can be explicitly written as:
\begin{align*}
    \underline{T}^i &= \sum_{p=1}^{d_0} \underline{\alpha}_{pi}+\sum_{q=1}^{d_0} \underline{\beta}_{qi}+b^{(\gate)}_i \\
    \overline{T}^i &= \sum_{p=1}^{d_0} \overline{\alpha}_{pi}+\sum_{q=1}^{d_0} \overline{\beta}_{qi}+b^{(\gate)}_i
\end{align*}
with $\underline{\alpha}_{pi}$, $\overline{\alpha}_{pi}$, $\underline{\beta}_{qi}$ and $\overline{\beta}_{qi}$ given by:
\begin{align*}
    \underline{\alpha}_{pi} &= \begin{cases} (\Theta^{(\gate)})_{pi}.\underline{v}_{w_t}^{p}, & (\Theta^{(\gate)})_{pi} \geq 0 \\
(\Theta^{(\gate)})_{pi}.\overline{v}_{w_t}^{p}, & (\Theta^{(\gate)})_{pi} < 0 
    \end{cases} \\
    \overline{\alpha}_{pi} &= \begin{cases} (\Theta^{(\gate)})_{pi}.\overline{v}_{w_t}^{p}, & (\Theta^{(\gate)})_{pi} \geq 0 \\
(\Theta^{(\gate)})_{pi}.\underline{v}_{w_t}^{p}, & (\Theta^{(\gate)})_{pi} < 0 
    \end{cases} \\
    \underline{\beta}_{qi} &= \begin{cases} (U^{(\gate)})_{qi}.\underline{h}_{t-1}^{q}, & (U^{(\gate)})_{qi} \geq 0 \\
    (U^{(\gate)})_{qi}.\overline{h}_{t-1}^{q}, & (U^{(\gate)})_{qi} < 0 
    \end{cases} \\
    \overline{\beta}_{qi} &= \begin{cases} (U^{(\gate)})_{qi}.\overline{h}_{t-1}^{q}, & (U^{(\gate)})_{qi} \geq 0 \\
    (U^{(\gate)})_{qi}.\underline{h}_{t-1}^{q}, & (U^{(\gate)})_{qi} < 0. 
    \end{cases}
\end{align*}
\\

The following proposition is adapted from \cite{meyer2021interval}.

\paragraph{Proposition 7:} Consider the map: 
$$
\begin{array}{l}
c_t : \quad\mathbf{V} \times \mathbf{H} \times \mathbf{C} \to \mathbb{R}^d \\
\quad\quad(v_{w_t},h_{t-1},c_{t-1}) \mapsto c_t(v_{w_t},h_{t-1},c_{t-1})
\end{array}
$$
defined in \eqref{eq:ct}, with $\mathbf{V}=[\underline{\mathbf{V}}, \overline{\mathbf{V}}]$, $\mathbf{H}=[\underline{\mathbf{H}}, \overline{\mathbf{H}}]$ and $\mathbf{C}=[\underline{\mathbf{C}}, \overline{\mathbf{C}}]$. Let $A^c =\dfrac{\partial c_t}{\partial v_{w_t}}\in \mathbb{R}^{d \times d_0}$, $B^c=\dfrac{\partial c_t}{\partial h_{t-1}} \in \mathbb{R}^{d \times d}$ and $D^c=\dfrac{\partial c_t}{\partial c_{t-1}} \in \mathbb{R}^{d \times d}$. \\
For all $i \in \{1,\ldots,d\}$, we have $c^i_t\in [\underline{c}^i_t,\overline{c}^i_t]$, with 
\begin{align*}
\underline{c}^i_t =& c^i_t\left(\frac{\overline{\mathbf{V}}+\underline{\mathbf{V}}}{2},\frac{\overline{\mathbf{H}}+\underline{\mathbf{H}}}{2},\frac{\overline{\mathbf{C}}+\underline{\mathbf{C}}}{2}\right)\\&-(A^c)_{i,:}(\overline{\mathbf{V}}-\underline{\mathbf{V}})-(B^c)_{i,:}(\overline{\mathbf{H}}-\underline{\mathbf{H}})\\&-(D^c)_{i,:}(\overline{\mathbf{C}}-\underline{\mathbf{C}})
\end{align*}
\begin{align*}
\overline{c}^i_t =&c^i_t\left(\frac{\overline{\mathbf{V}}+\underline{\mathbf{V}}}{2},\frac{\overline{\mathbf{H}}+\underline{\mathbf{H}}}{2},\frac{\overline{\mathbf{C}}+\underline{\mathbf{C}}}{2}\right)\\&+(A^c)_{i,:}(\overline{\mathbf{V}}-\underline{\mathbf{V}})+(B^c)_{i,:}(\overline{\mathbf{H}}-\underline{\mathbf{H}})\\&+(D^c)_{i,:}(\overline{\mathbf{C}}-\underline{\mathbf{C}})
\end{align*}
where $(A^c)_{i,:}$ (respectively $(B^c)_{i,:}$ and $(D^c)_{i,:}$) denotes the i-th row vector of the matrix $A^c$ (respectively $B^c$ and $D^c$). 

\section{Proofs of Model-Specific GBM}
\label{sec:proofs}

In this section, we present the proofs for the main results of this paper, specifically Proposition 2, Proposition 3, and Proposition 4. For S4 and CNN architectures, the GBM is directly derived from the learned parameters, making its computation straightforward. In contrast, for LSTM models, the GBM computation is more involved, requiring bounds on specific functions. To address this issue, we provide a detailed algorithmic procedure for computing the GBM in the LSTM case, ensuring a practical and efficient implementation.  

\subsection{Proof of Proposition 2}
Consider the map $\mathcal{F}$ describing the input-output model on an LSTM cell defined in (\ref{eq:lstm}), where the input is \(\ x=(v_{w_t}, h_{t-1}, c_{t-1}) \in \mathbf{V} \times \mathbf{H} \times \mathbf{C} \subseteq \mathbb{R}^{d_0+2d}\) and the output is \(\,y_t = \mathcal{F}(v_{w_t}, h_{t-1}, c_{t-1}) = o_t \odot \tanh(f_t \odot c_{t-1} + I_t \odot g_t) \in \mathbb{R}^d\), with,\\
\begin{align}
    I_t &= \sigma\left(\Theta^{(I)}. v_{w_t} + U^{(I)} .h_{t-1} + b^{(I)}\right)  \label{eq:it}\\
    f_t &= \sigma\left(\Theta^{(f)} .v_{w_t} + U^{(f)} .h_{t-1} + b^{(f)}\right)  \label{eq:ft}\\
    g_t &= \tanh\left(\Theta^{(g)} .v_{w_t} + .U^{(g)} h_{t-1} + b^{(g)}\right)  \label{eq:gt}\\
    o_t &= \sigma\left(\Theta^{(o)} .v_{w_t} + U^{(o)} .h_{t-1} + b^{(o)}\right)
     \label{eq:ot}
\end{align}where $\sigma$ denotes the sigmoid function and $\tanh$ denotes the hyperbolic tangent function. For $\gate \in \{I,f,g,o\}$, the parameters $\Theta^{(gate)} \in \mathbb{R}^{d\times d_0} , U^{(gate)} \in \mathbb{R}^{d\times d}$ and $b^{(gate)} \in \mathbb{R}^{d}$ are the input-hidden weights, hidden-hidden weights, and biases, respectively.\\

Lets consider the following:
\begin{small}
    \begin{equation*}
    T_{\gate}^i = \sum_{p=1}^{d_0} (\Theta^{(\gate)})_{ip}.v_{w_t}^{p} + \sum_{q=1}^{d} (U^{(\gate)})_{iq}.h_{t-1}^{q} + b_i^{(\gate)}
\end{equation*}
\end{small}
with \(\sigma'\) and \(\tanh'\) are the derivative of the sigmoid and hyperbolic tangent functions, respectively.

Here, \(\mathbf{V}\) denotes the domain of the word vectors, \(\mathbf{H}\) denotes the domain of hidden states, and \(\mathbf{C}\) denotes the domain of cell states. The i-th component of the map $\mathcal{F}$ is given for $i \in \{1,\ldots,d\}$ by: 
$$\mathcal{F}^i(x)=o_t^i \odot \tanh(f^i_t \odot c^i_{t-1} + I^i_t \odot g^i_t). $$ 
Now consider $j \in \{1,\ldots,d_0+2d\}$ and consider the j-th component of the input vector $x \in \mathbf{V} \times \mathbf{H} \times \mathbf{C} \subseteq \mathbb{R}^{d_0+2d}$. Since $x$ is the concatenation of the vectors $v_{w_t} \in \mathbf{V}\subseteq \mathbb{R}^{d_0}$ and $h_{t-1}\in\mathbf{H}\subseteq \mathbb{R}^{d}$, and $c_{t-1}\in\mathbf{C}\subseteq \mathbb{R}^{d}$, we distinguish three cases: if $j \in \{1,\ldots,d_0\}$, then $x^j$ is a component of the vector $v_{w_t}$ and one gets
\begin{align*}
     \dfrac{\partial \mathcal{F}^i}{\partial x^j}(x) &= (\mathcal{M}_v)_{i,j}(x)\\ &=\dfrac{\partial o^i_t}{\partial x^j} \cdot \tanh(c^i_t) + o^i_t \cdot \dfrac{\partial \tanh(c^i_t)}{\partial x^j} 
\end{align*}
Using Eq. \eqref{eq:ot}, one gets \[\dfrac{\partial o_t^i}{\partial x^j} = (\Theta^{(o)})_{i,j}\cdot \sigma'(T_{o}^i)\] and \(\dfrac{\partial \tanh(c^i_t)}{\partial x^j} = \dfrac{\partial c^i_t}{\partial x^j}\cdot \tanh'(c^i_t)\), 

with \(\sigma'\) and \(\tanh'\) are the derivative of the sigmoid and hyperbolic tangent functions, respectively.\\
Based on Eqs.(\ref{eq:ct}),\eqref{eq:it},\eqref{eq:ft} and \eqref{eq:gt}, we have the following:
\begin{align*}
    \frac{\partial c^i_t}{\partial x^j} &= (\Theta^{(f)})_{i,j} \cdot \sigma'(T_f^i) \cdot c^i_{t-1} \\
    &+ (\Theta^{(I)})_{i,j} \cdot\sigma'(T_I^i) \cdot\tanh(T_g^i) \\
    &+ (\Theta^{(g)})_{i,j}\cdot \sigma(T_I^i)\cdot\tanh'(T_g^i).
\end{align*}
which implies that
\begin{align*}
     (\mathcal{M}_v)_{i,j}(x) &= (\Theta^{(o)})_{i,j} .\sigma'(T_o^i) .\tanh(c^i_{t}) \\
    &+ \sigma(T_o^i) .\frac{\partial c^i_t}{\partial x^j} .\tanh'(c^i_{t}) 
\end{align*}
Hence, it follows that
$$\left\lVert \dfrac{\partial \mathcal{F}^i}{\partial x^j}(x)\right\rVert \leq  \max\left(\| (\underline{\mathcal{M}_v})_{ij} \|, \| (\overline{\mathcal{M}_v})_{ij} \|\right).$$

Similarly, if $j \in \{d_0+1,\ldots,d_0+d\}$ then $x^j$ is a component of the vector $h_{{t-1}}$ and one gets
\begin{align*}
     \dfrac{\partial \mathcal{F}^i}{\partial x^j}(x) &= (\mathcal{M}_h)_{i,j-d_0}(x)\\ &=\dfrac{\partial o^i_t}{\partial x^j} \cdot \tanh(c^i_t) + o^i_t \cdot \dfrac{\partial \tanh(c^i_t)}{\partial x^j} 
\end{align*}
Using Eq. \eqref{eq:ot}, one gets \[\dfrac{\partial o_t^i}{\partial x^j} = (U^{(o)})_{i,j}\cdot \sigma'(T_{o}^i)\] and \(\dfrac{\partial \tanh(c^i_t)}{\partial x^j} = \dfrac{\partial c^i_t}{\partial x^j}\cdot \tanh'(c^i_t)\), \\

Based on Eqs.(\ref{eq:ct}), \eqref{eq:it}, \eqref{eq:ft} and \eqref{eq:gt}, we have the following:
\begin{align*}
    \frac{\partial c^i_t}{\partial x^j} &= (U^{(f)})_{i,j} \cdot \sigma'(T_f^i) \cdot c^i_{t-1} \\
    &+ (U^{(I)})_{i,j} \cdot\sigma'(T_I^i) \cdot\tanh(T_g^i) \\
    &+ (U^{(g)})_{i,j}\cdot \sigma(T_I^i)\cdot\tanh'(T_g^i).
\end{align*}
which implies that:
\begin{align*}
     (\mathcal{M}_h)_{i,j-d_0}(x) &= (U^{(o)})_{i,j} .\sigma'(T_o^i) .\tanh(c^i_{t}) \\
    &+ \sigma(T_o^i) .\frac{\partial c^i_t}{\partial x^j} .\tanh'(c^i_{t}) 
\end{align*} 
Hence, it follows that

$$\left\lVert \dfrac{\partial \mathcal{F}^i}{\partial x^j}(x)\right\rVert \leq  \max\left(\| (\underline{\mathcal{M}_h})_{i,j-d_0} \|, \| (\overline{\mathcal{M}_h})_{i,j-d_0} \|\right).$$

Finally, if $j \in \{d_0+d+1,\ldots,d_0+2d\}$ then $x^j$ is a component of the vector $c_{{t-1}}$ and one gets
\begin{align*}
     \dfrac{\partial \mathcal{F}^i}{\partial x^j}(x) &= (\mathcal{M}_c)_{i,j-d_0-d}(x)\\ 
     &= \dfrac{\partial o^i_t}{\partial x^j} \cdot \tanh(c^i_t) + o^i_t \cdot \dfrac{\partial \tanh(c^i_t)}{\partial x^j}
\end{align*}
Based on Eqs.(\ref{eq:ct}), one get:
\begin{align*}
     \dfrac{\partial \mathcal{F}^i}{\partial x^j}(x) &= (\mathcal{M}_c)_{i,j-d_0-d}(x)\\  &=\sigma(T_o^i).\sigma(T_f^i).\tanh'(c^i_{t})
\end{align*}
Hence, it follows that
$\left\lVert \dfrac{\partial \mathcal{F}^i}{\partial x^j}(x)\right\rVert \leq  \max\left(\| (\underline{\mathcal{M}_c})_{i,j-d_0-d} \|, \| (\overline{\mathcal{M}_c})_{i,j-d_0-d} \|\right)$.\hfill $\square$
\\

The computation of the Growth Bound Matrix (GBM) for LSTM models follows a systematic procedure grounded in analytical bounds and algorithmic implementations. For each component pair \((i, j)\), the process begins by bounding the intermediate term \(T_o^i\) using \textit{Proposition~6}. Subsequently, the derivative of the sigmoid function \(\sigma'(T_o^i)\) is bounded using \textit{Proposition~5}, and the cell state \(c_t^i\) is bounded via \textit{Proposition~7}. The derivative \(\tanh'(c_t^i)\) is then bounded again using \textit{Proposition~5}. These bounds form the basis for the estimation of partial derivatives of the cell state with respect to both the input embedding \(v_{w_t}^j\) and the hidden state \(h_{t-1}^j\), which are computed using \textit{Algorithm~\ref{alg:algo4}} and \textit{Algorithm~\ref{alg:algo5}}, respectively.

The complete bounds of the GBM components are derived through three dedicated algorithms:
\begin{itemize}
    \item \textbf{Algorithm~\ref{alg:algo1}} computes the upper and lower bounds of the matrix \(\mathcal{M}_v\), corresponding to the sensitivity of the hidden state with respect to the input embedding \(v_{w_t}\). This computation integrates the intermediate bounds and uses \textit{Algorithm~\ref{alg:algo4}} for estimating \(\partial c_t^i / \partial v_{w_t}^j\).
    \item \textbf{Algorithm~\ref{alg:algo2}} estimates the boundaries of \(\mathcal{M}_h\), which represents the sensitivity of the hidden state with respect to the previous hidden state \(h_{t-1}\). The estimation is based on Propositions~5, 6, and 7, and it incorporates \textit{Algorithm~\ref{alg:algo5}} to bound \(\partial c_t^i / \partial h_{t-1}^j\).
    
    \item \textbf{Algorithm~\ref{alg:algo3}} computes the bounds of \(\mathcal{M}_c\), corresponding to the impact of the previous cell state \(c_{t-1}\) on the current hidden state. This step uses the bounds on \(T_o^i\), \(T_f^i\), and \(c_t^i\), and again relies on the propositions 5, 6, and 7.
\end{itemize}

Each matrix component is ultimately characterized by a pair of upper and lower bounds, denoted \((\overline{\mathcal{M}_v})_{i,j}, (\underline{\mathcal{M}_v})_{i,j}\), \((\overline{\mathcal{M}_h})_{i,j}, (\underline{\mathcal{M}_h})_{i,j}\), and \((\overline{\mathcal{M}_c})_{i,j}, (\underline{\mathcal{M}_c})_{i,j}\), respectively. These are assembled to construct the final GBM matrices \(\mathcal{M}_v\), \(\mathcal{M}_h\), and \(\mathcal{M}_c\), as defined in Equation~\eqref{eq:GBMLSTM}. The overall GBM \(\mathcal{M}\) is then formed according to the formulation detailed in \textit{Proposition 2 (Sec. \ref{prop2})}.

\begin{figure*}[ht]
    \centering
    \includegraphics[width=16cm]{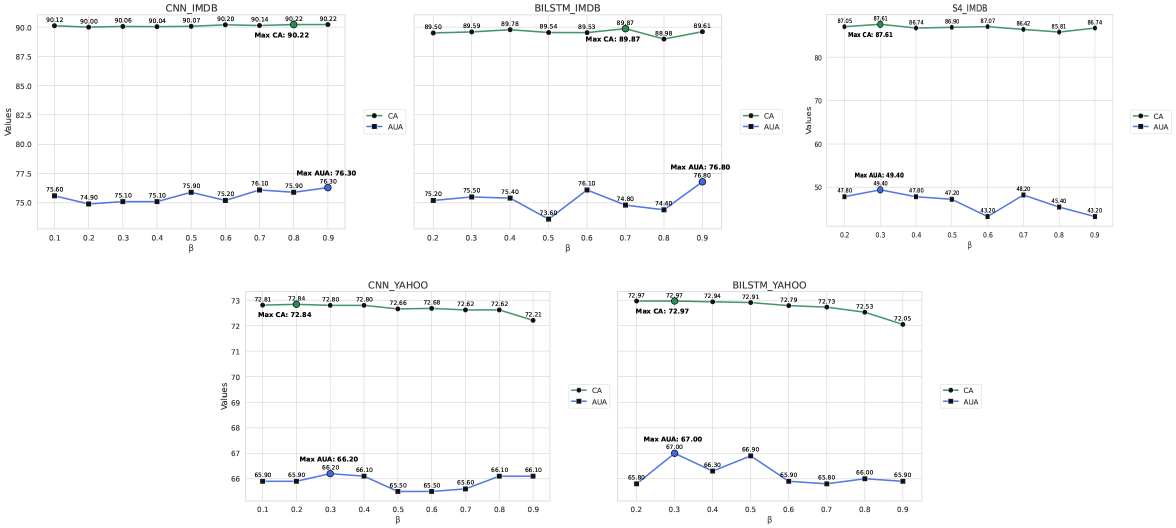}
    \caption{The impact of the hyperparameter $\beta$ on the performance of CNN, BiLSTM, and S4 models when subjected to the PWWS adversarial attack. The evaluation is conducted on the \textit{Yahoo! Answers} and \textit{IMDB} datasets, with CNN and BiLSTM assessed on both datasets, while S4 is evaluated exclusively on \textit{IMDB}.}
    \label{fig:fig3}
\end{figure*}

\subsection{Proof of Proposition 3}
Consider the map $\mathcal{F}$ describing the input-output model on an S4 cell defined in (\ref{eqn:SSM}), where the input is \(x = (v_{w_t}, h_{t-1})\in\mathbb{R}^{d_0+d_0d}\) and the output is \(y_t= \mathcal{F}(v_{w_t}, h_{t-1})=\tilde{C}\bigl(\tilde{A}h_{t-1} + \tilde{B}v_{w_t}\bigr) + \tilde{D}v_{w_t} \in \mathbb{R}^{d_0}\). The i-th component of the map $\mathcal{F}$ is given for $i \in \{1,\ldots,d_0\}$ by 
$$\mathcal{F}^i(x)=(\tilde{C}\tilde{A})_{i,:}h_{t-1} + ((\tilde{C}\tilde{B})_{i,:}+ (\tilde{D})_{i,:})v_{w_t} $$
where the notation $(\mathcal{M})_{i,:}$ is used to denote the i-th row of the matrix $\mathcal{M}$. Now consider $j \in \{1,\ldots,d_0+d_0d\}$ and consider the j-th component of the input vector $x \in \mathbb{R}^{d_0+d_0d}$. Since $x$ is the concatenation of the vectors $v_{w_t} \in \mathbb{R}^{d_0}$ and $h_{t-1}\in\mathbb{R}^{d_0d}$, we distinguish two cases: if $j \in \{1,\ldots,d_0\}$, then $x^j$ is a component of the vector $v_{w_t}$ and one gets
\begin{align*}
     \left\lVert \dfrac{\partial \mathcal{F}^i}{\partial x^j}(x)\right\rVert \leq \| (\tilde{C}\tilde{B})_{i,j}+ (\tilde{D})_{i,j}\|. 
\end{align*}
Similarly, if $j \in \{d_0+1,\ldots,d_0+d_0d\}$ then $x^j$ is a component of the vector $h_{{t-1}}$ and one gets
\begin{align*}
     \left\lVert \dfrac{\partial \mathcal{F}^i}{\partial x^j}(x)\right\rVert \leq \| (\tilde{C}\tilde{A})_{i,j-d_0}\|
     \end{align*}
     which concludes the proof. \hfill $\square$

{\subsection{Proof of Proposition~4}}

Consider the map \(\mathcal{F}\) defined in Eq.~\eqref{eq:cnn}, which characterizes the input-output transformation of a convolutional neural network (CNN). The input is a flattened embedding vector \(x = v_x \in \mathbb{R}^{N d_0}\), and the output is obtained by applying convolution and pooling operations across multiple kernel sizes. Specifically, the output is given by:
\[
y = \bigoplus_{k \in \mathcal{K}} \mathcal{P}^{(k)} \circ \mathcal{C}^{(k)}(v_x) \in \mathbb{R}^{|\mathcal{K}| \cdot d},
\]
where \(\mathcal{K} = \{k_1, \dots, k_m\}\) denotes the set of kernel sizes, and \(d\) is the number of output channels (filters) per kernel size.

Let us now describe how the individual components of \(\mathcal{F}\) are computed. For a given kernel size \(k_p \in \mathcal{K}\), the output of the corresponding convolution and max-pooling operation is a \(d\)-dimensional vector. The \(i\)-th component of this vector is given by:
\[
\mathcal{F}^i_p(x) = \max_{t = 1}^{N - k_p + 1} \phi\left(b_i^{(k_p)} + \sum_{l = 0}^{k_p - 1} W_{i,:,l}^{(k_p)} \cdot v_{w_{t + l}}\right),
\]
where:
\begin{itemize}
    \item \(W^{(k_p)} \in \mathbb{R}^{d \times d_0 \times k_p}\) is the convolutional kernel for size \(k_p\),
    \item \(b_i^{(k_p)}\) is the bias term for the \(i\)-th filter,
    \item \(\phi\) is the activation function (e.g., ReLU or tanh),
    \item and the max operator is applied across the temporal dimension \(t\).
\end{itemize}

After processing all kernel sizes in \(\mathcal{K}\), the final output vector \(y = \mathcal{F}(x)\) is constructed by concatenating the outputs from each kernel size. Thus, each output index \(i \in \{1, \dots, |\mathcal{K}| \cdot d\}\) corresponds to a specific kernel and a specific filter within that kernel.

To identify which kernel and filter an output index \(i\) corresponds to, we define two functions:
\[
\alpha(i, a, d) = \left\lfloor \frac{i - a}{d} \right\rfloor + 1, \]
\[
\beta(i, a, d) = 1 + ((i - a) \bmod d),
\]
where:
\begin{itemize}
    \item \(\alpha(i,1,d)\) gives the index of the kernel \(k_{\alpha(i,1,d)}\) associated with output index \(i\),
    \item \(\beta(i,1,d)\) gives the index of the filter within the kernel.
\end{itemize}

Thus, the \(i\)-th output of \(\mathcal{F}(x)\) can be written as:
\[
\mathcal{F}^i(x) = \max_{t = 1}^{N - k_{\alpha(i,1,d)} + 1} \phi\left(\mathcal{Q}_i\right),
\]
where:
\[
\mathcal{Q}_i = b^{(k_{\alpha(i,1,d)})}_i + \sum_{l = 0}^{k_{\alpha(i,1,d)} - 1} W^{(k_{\alpha(i,1,d)})}_{\beta(i,1,d), :, l} \cdot v_{w_{t + l}}.
\]

We now analyze the GBM of the output \(\mathcal{F}^i(x)\) with respect to a component \(x^j\) of the input. Since the Lipschitz constants of the used activations function (ReLU or Tanh) are bounded by $1$, the derivative of \(\mathcal{F}^i\) with respect to \(x^j\) is bounded by the magnitude of the associated weight in the convolution kernel.

Let \(j \in \{1, \dots, N d_0\}\) denote a specific input index. We identify the relevant input slice using:
\[
\alpha(j,t,d_0) = \left\lfloor \frac{j - t}{d_0} \right\rfloor + 1, \] 
\[\beta(j,t,d_0) = 1 + ((j - t) \bmod d_0).
\]

Then, the partial derivative of \(\mathcal{F}^i\) with respect to \(x^j\) satisfies:
\[
\left\lVert \frac{\partial \mathcal{F}^i}{\partial x^j}(x) \right\rVert \leq \max_{t = 1}^{j} \left\lVert W^{(k_{\alpha(i,1,d)})}_{\beta(i,1,d), \beta(j,t,d_0), \alpha(j,t,d_0)} \right\rVert.
\]

Taking the supremum over all valid \(t\), this yields the expression for the GBM as stated in Eq.~\eqref{eq:GBMcnn}, which completes the proof. \hfill $\square$

\newpage
\section{Experiment Details}
\label{sec:Exp}

\subsection{Additional experiments: Clean Accuracy versus Robust Accuracy}

\paragraph{Hyper-parameter Study:} In Eq.\eqref{eq:objective}, the loss function of the model incorporates a hyper-parameter $\beta$ to regulate the trade-off between clean accuracy and robustness accuracy. This trade-off varies depending on the model, as CNN, BiLSTM, and S4 demonstrate different sensitivity levels on the Yahoo! Answers and IMDB datasets as shown in Figure \ref{fig:fig3}. These results underscore the challenge of maintaining both robustness and accuracy in adversarial settings, highlighting the importance of carefully tuning $\beta$ to optimize performance for each model and dataset.

\subsection{Dataset Statistics}
\begin{table}[ht]
    \centering
    \begin{tabular}{@{}lccc@{}}
        \toprule
         Dataset & Training set & Test set & Classes \\
         \midrule
         IMDB & 25,000 & 25,000 & 2 \\
         Yahoo! Answers & 1,400,000 & 50,000 & 10 \\
        \bottomrule
    \end{tabular}
    \caption{Statistics of the \textit{IMDB} and \textit{Yahoo! Answers} datasets.}
    \label{tab:stat}
\end{table}

\subsection{Embedding layer}
In our experiments, we freeze the parameters of the pre-trained embedding layer and update only those of the classification model.

\subsection{Detailed Setup}

In the BiLSTM layer, the weights $\Theta^{(\text{gate})}$ and $U^{(\text{gate})}$ are shared between the cells. For classification tasks, the final hidden state $h_N$ is typically considered. Therefore, when calculating the GBM, we focus solely on the last cell in both the forward and backward passes. This calculation is performed using parallel computing to expedite the process.
The details of the GBM training method for BiLSTM model are provided in Table \ref{tab:conf}.

\begin{table}[ht]
    \centering
    \begin{tabular}{lc|c}
    \hline
    \textbf{Dataset} & IMDB & Yahoo! Answers \\
    \hline \hline
    \textbf{Optimizer} & \multicolumn{2}{c}{$\operatorname{Adam}$} \\
    \hline
    \textbf{Batch size} & \multicolumn{2}{c}{64} \\
    \hline
    \textbf{Hidden size} & \multicolumn{2}{c}{64} \\
    \hline
    \textbf{Learning rate} & \multicolumn{2}{c}{$10^{-3}$} \\
    \hline
    \textbf{Weight decay} & \multicolumn{2}{c}{$10^{-4}$} \\
    \hline
    \textbf{Max length} & 512 & 256 \\
    \hline
    \textbf{Early Stopping} & \multicolumn{2}{c}{Yes} \\
    \hline
    \end{tabular}
    \caption{Training configuration and hyperparameters of the GBM training method for BiLSTM model.}
    \label{tab:conf}
\end{table}

\hspace{-4mm}In S4 layer, the parameters \(A, B, C,\) and $\Delta$ are typically require a smaller learning rate, with no weight decay. The Table \ref{tab:conf_S4} outlines the GBM training procedure for the S4 model.

\begin{table}[ht]
    \centering
    \resizebox{\columnwidth}{!}{
    \begin{tabular}{lc|c}
    \hline
    \textbf{Parameters} & \(A, B, C,\Delta\) & Others \\
    \hline \hline
    \textbf{Optimizer} & \multicolumn{2}{c}{$\operatorname{Adam}$} \\
    \hline
    \textbf{Batch size} & \multicolumn{2}{c}{64} \\
    \hline
    \textbf{Hidden size} & \multicolumn{2}{c}{256} \\
    \hline
    \textbf{Learning rate} & {$5.10^{-3}$} & {$5.10^{-4}$}  \\
    \hline
    \textbf{Weight decay} & 0 & {$10^{-2}$} \\
    \hline
    \textbf{Max length} & \multicolumn{2}{c}{512}  \\
    \hline
    \textbf{Early Stopping} & \multicolumn{2}{c}{Yes} \\
    \hline
    \end{tabular}}
    \caption{Overview of GBM Training Setup and Hyperparameters for S4 Model on the \textit{IMDB} dataset.}
    \label{tab:conf_S4}
\end{table}

\hspace{-4mm}The CNN model is trained using the GBM approach as describe the Table \ref{tab:conf_cnn}.
All experiments were conducted on a single NVIDIA A100 80GB GPU. 
\begin{table}[H]
    \centering
    \resizebox{\columnwidth}{!}{
    \begin{tabular}{lc|c}
    \hline
    \textbf{Dataset} & IMDB & Yahoo! Answers \\
    \hline \hline
    \textbf{Optimizer} & \multicolumn{2}{c}{$\operatorname{Adam}$} \\
    \hline
    \textbf{Batch size} & \multicolumn{2}{c}{64} \\
    \hline
    \textbf{Num kernels} & \multicolumn{2}{c}{128} \\
    \hline
    \textbf{Learning rate} & \multicolumn{2}{c}{$10^{-4}$} \\
    \hline
    \textbf{Weight decay} & \multicolumn{2}{c}{$10^{-4}$} \\
    \hline
    \textbf{Kernel sizes} & \multicolumn{2}{c}{(3, 4, 5)} \\
    \hline
    \textbf{Max length} & 512 & 256 \\
    \hline
    \textbf{Early Stopping} & \multicolumn{2}{c}{Yes} \\
    \hline
    \end{tabular}}
    \caption{Hyperparameter Settings and Training Configuration for GBM in CNN Model.}
    \label{tab:conf_cnn}
\end{table}
\section{Adversarial examples}
The Figure \ref{fig:samples} demonstrates a synonym substitution attack, in which minor word replacements alter the model's sentiment classification while maintaining the original meaning of the text.
\begin{figure}[H]
    \centering
    \includegraphics[width=7.5cm]{ 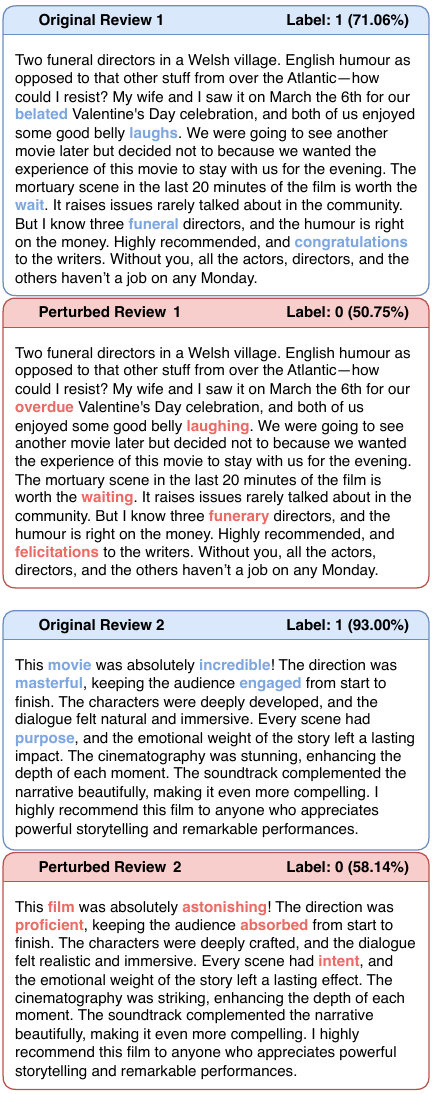}
    \caption{Examples of a synonym substitution attack, where words from the original review (\textbf{blue}) are replaced with their synonyms (\textbf{red}), preserving the semantic meaning of the text but resulting in sentiment misclassification. For instance, in Review 1, replacing words such as "belated" with "overdue", "belly laughs" with "belly laughing", and "congratulations" with "felicitations" leads the prediction to shift from positive (Label: 1, 71.06\%) to negative (Label: 0, 50.75\%). Similarly, in Review 2, substituting "incredible" with "astonishing", "masterful" with "proficient", and "developed" with "crafted" results in a classification flip from positive (Label: 1, 93\%) to negative (Label: 0, 58.14\%).}
    \label{fig:samples}
\end{figure}

\section{Algorithms}
\label{sec:Algo}

\begin{algorithm}[ht]
        \caption{The boundaries of \(\mathcal{M}_v\)}
        \label{alg:algo1}
        \KwIn{$\Theta$, $U$: Weights; $b$: bias; $\mathbf{V}$, $\mathbf{H}$, $\mathbf{C}$: input intervals}
        \KwOut{The boundaries of \(\mathcal{M}_v\)}
        \BlankLine
        
        \For{$i \gets 1$ \KwTo $d$}{
            \For{$j \gets 1$ \KwTo $d_{0}$}{
                1. {Bounding $T_o^i$ by \textit{Proposition 6} (Appendix \ref{sec:AR})}\;
                2. {Bounding $\sigma'(T_o^i)$ by \textit{Proposition 5}} (Appendix \ref{sec:AR})\;
                3. {Bounding $c_t^i$ by \textit{Proposition 7} (Appendix \ref{sec:AR})}\;
                4. {Bounding $\tanh'(c_t^i)$ by \textit{Proposition 5} (Appendix \ref{sec:AR})}\;
                5. {Bounding \(\frac{\partial c^i_t}{\partial v_{w_t}^j}\) by \textit{Algorithm.\ref{alg:algo4}} (Appendix \ref{sec:Algo})}\;
                
                $[(\underline{\mathcal{M}_v})_{i,j}, (\overline{\mathcal{M}_v})_{i,j}] \gets$ The boundaries of $\mathcal{M}_v$\;
            }
        }
        
\end{algorithm}

\begin{algorithm}[htbp]
        \caption{The boundaries of \(\mathcal{M}_h\)}
        \label{alg:algo2}
        \KwIn{$\Theta$, $U$: Weights; $b$: bias; $\mathbf{V}$, $\mathbf{H}$, $\mathbf{C}$: input intervals}
        \KwOut{The boundaries of \(\mathcal{M}_h\)}
        \BlankLine
        
        \For{$i \gets 1$ \KwTo $d$}{
            \For{$j \gets 1$ \KwTo $d$}{
                1. Bounding $T_o^i$ by \textit{Proposition 6} (Appendix \ref{sec:AR})\;
                2. Bounding $\sigma'(T_o^i)$ by \textit{Proposition 5} (Appendix \ref{sec:AR})\;
                3. Bounding $c_t^i$ by \textit{Proposition 7} (Appendix \ref{sec:AR})\;
                4. Bounding $\tanh'(c_t^i)$ by \textit{Proposition 5} (Appendix \ref{sec:AR})\;
                5. Bounding \(\frac{\partial c^i_t}{\partial h_{t-1}^j}\) by \textit{Algorithm.\ref{alg:algo5}}\;
                
                $[(\underline{\mathcal{M}_h})_{i,j}, (\overline{\mathcal{M}_h})_{i,j}] \gets$ The boundaries of $\mathcal{M}_h$\;
            }
        }
        
\end{algorithm}

\begin{algorithm}[htbp]
       \caption{The boundaries of \(\mathcal{M}_c\)}
       \label{alg:algo3}
       \KwIn{$\Theta$, $U$: Weights; $b$: bias; $\mathbf{V}$, $\mathbf{H}$, $\mathbf{C}$: input intervals}
       \KwOut{The boundaries of \(\mathcal{M}_c\)}
       \BlankLine
        
       \For{$i \gets 1$ \KwTo $d$}{
           \For{$j \gets 1$ \KwTo $d$}{
               1. Bounding $T_o^i$ and $T_f^i$ by \textit{Proposition 6} (Appendix \ref{sec:AR})\;
               2. Bounding $c_t^i$ by \textit{Proposition 7} (Appendix \ref{sec:AR})\;
               3. Bounding $\tanh'(c_t^i)$ by \textit{Proposition 5} (Appendix \ref{sec:AR})\;
                
               $[(\underline{\mathcal{M}_c})_{i,j}, (\overline{\mathcal{M}_c})_{i,j}] \gets$ The boundaries of $\mathcal{M}_c$\;
           }
       }
        
\end{algorithm}

\begin{algorithm}[htbp]
        \caption{The boundaries of $A^c$}
        \label{alg:algo4}
        \KwIn{$\Theta$, $U$: Weights; $b$: bias; $\mathbf{V}$, $\mathbf{H}$, $\mathbf{C}$: input intervals}
        \KwOut{The boundaries of $A^c$}
        \BlankLine
        
        \For{$i \gets 1$ \KwTo $d$}{
            \For{$j \gets 1$ \KwTo $d$}{
                1. Bounding $T_f^i$, $T_I^i$ and $T_g^i$ by \textit{Proposition 6} (Appendix \ref{sec:AR}) \;
                2. Bounding $\sigma'(T_f^i)$, $\sigma'(T_I^i)$ and $\sigma'(T_g^i)$ by \textit{Proposition 5} (Appendix \ref{sec:AR})\;
                3. Bounding $c_t^i$ by \textit{Proposition 7} (Appendix \ref{sec:AR}) \;
                4. Using the monotonicity proprety of \(\sigma\) and \(\tanh\) functions\;
                
                $[(\underline{A^c})_{i,j}, (\bar{A^c})_{i,j}] \gets$ The boundaries of $A^c$\;
            }
        }
        
\end{algorithm}

\begin{algorithm}[htbp]
        \caption{The boundaries of $B^c$}
        \label{alg:algo5}
        \KwIn{$\Theta$, $U$: Weights; $b$: bias; $\mathbf{V}$, $\mathbf{H}$, $\mathbf{C}$: input intervals}
        \KwOut{The boundaries of $B^c$}
        \BlankLine
        
        \For{$i \gets 1$ \KwTo $d$}{
            \For{$j \gets 1$ \KwTo $d$}{
                1. Bounding $T_f^i$, $T_I^i$ and $T_g^i$ by \textit{Proposition 6} (Appendix \ref{sec:AR}) \;
                2. Bounding $\sigma'(T_f^i)$, $\sigma'(T_I^i)$ and $\sigma'(T_g^i)$ by \textit{Proposition 5} (Appendix \ref{sec:AR})\;
                3. Bounding $c_t^i$ by \textit{Proposition 7} (Appendix \ref{sec:AR}) \;
                4. Using the monotonicity proprety of \(\sigma\) and \(\tanh\) functions\;
                
                $[(\underline{B^c})_{ij}, (\bar{B^c})_{ij}] \gets$ The boundaries of $B^c$\;
            }
        }
        
\end{algorithm}


\end{document}